\documentclass[letterpaper, 10 pt, conference]{ieeeconf}  

\IEEEoverridecommandlockouts 

\usepackage{graphicx} 
\graphicspath{{figs/}}
\usepackage[fleqn]{amsmath} 
\usepackage{amssymb}  
\usepackage[fleqn]{mathtools}

\usepackage{multicol}
\usepackage{multirow}
\usepackage{placeins}
\usepackage{diagbox}
\usepackage{todonotes}
\usepackage{adjustbox}
\usepackage{verbatim}
\usepackage{cite}
\usepackage{wrapfig}
\usepackage{booktabs}
\usepackage{url}
\usepackage[utf8x]{inputenc}

\usepackage[noline,linesnumbered]{algorithm2e}
\SetArgSty{textnormal} 

\DeclareMathOperator{\argmax}{argmax}
\DeclareMathOperator{\argmin}{argmin}

\title{\LARGE Probabilistic approach to physical object disentangling}

\author{Joni Pajarinen$^{1,2}$, Oleg Arenz$^{1,3}$, Jan Peters$^{1,4}$, Gerhard Neumann$^{3}$ 
\thanks{This work was supported by EU H2020 project RoMaNS \#645582, ERC StG \#640554 (SKILLS4ROBOTS), and DFG project PA 3179/1-1 (ROBOLEAP)}%
\thanks{$^{1}$Intelligent Autonomous Systems, TU Darmstadt, Germany}
\thanks{$^{2}$Tampere University, Finland}
\thanks{$^{3}$Lincoln Center for Autonomous Systems, University of Lincoln, UK}
\thanks{$^{4}$MPI for Intelligent Systems, Tuebingen \newline
  {\tt\small \{pajarinen,arenz,peters\} @ ias.tu-darmstadt.de},
{\tt\small gneumann @ lincoln.ac.uk}}%
\thanks{\copyright 2020 IEEE.  Personal use of this material is permitted.  Permission from IEEE must be obtained for all other uses, in any current or future media, including reprinting/republishing this material for advertising or promotional purposes, creating new collective works, for resale or redistribution to servers or lists, or reuse of any copyrighted component of this work in other works.}%
}

\begin{document}

\maketitle
\thispagestyle{empty}
\pagestyle{empty}

\begin{abstract}
Physically disentangling entangled objects from each other is a
problem encountered in waste segregation or in any task that requires
disassembly of structures. Often there are no object models, and,
especially with cluttered irregularly shaped objects, the robot can
not create a model of the scene due to occlusion. One of our key
insights is that based on previous sensory input we are only
interested in moving an object out of the disentanglement around
obstacles. That is, we only need to know where the robot can
successfully move in order to plan the disentangling. Due to the
uncertainty we integrate information about blocked movements into a
probability map. The map defines the probability of the robot
successfully moving to a specific configuration. Using as cost the
failure probability of a sequence of movements we can then plan and
execute disentangling iteratively. Since our approach circumvents only
previously encountered obstacles, new movements will yield information
about unknown obstacles that block movement until the robot has
learned to circumvent all obstacles and disentangling succeeds. In the
experiments, we use a special probabilistic version of the Rapidly
exploring Random Tree (RRT) algorithm for planning and demonstrate
successful disentanglement of objects both in 2-D and 3-D simulation,
and, on a KUKA LBR 7-DOF robot. Moreover, our approach outperforms
baseline methods.
\end{abstract}

\section{Introduction}
\begin{figure}[t]
	\centering \vspace{0.8em}
	\hspace{-0.4em}\includegraphics[width=0.48\textwidth]{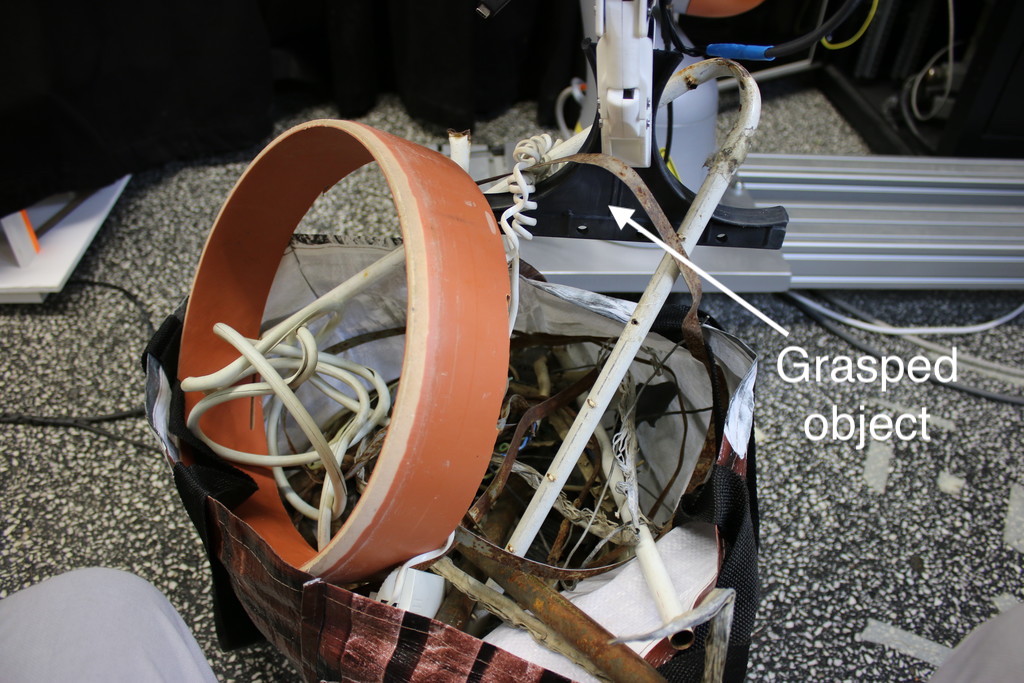}\\
	\vspace{2em}
	\begin{tabular}{cccc}
		\hspace{-0.5em}\includegraphics[width=0.117\textwidth]{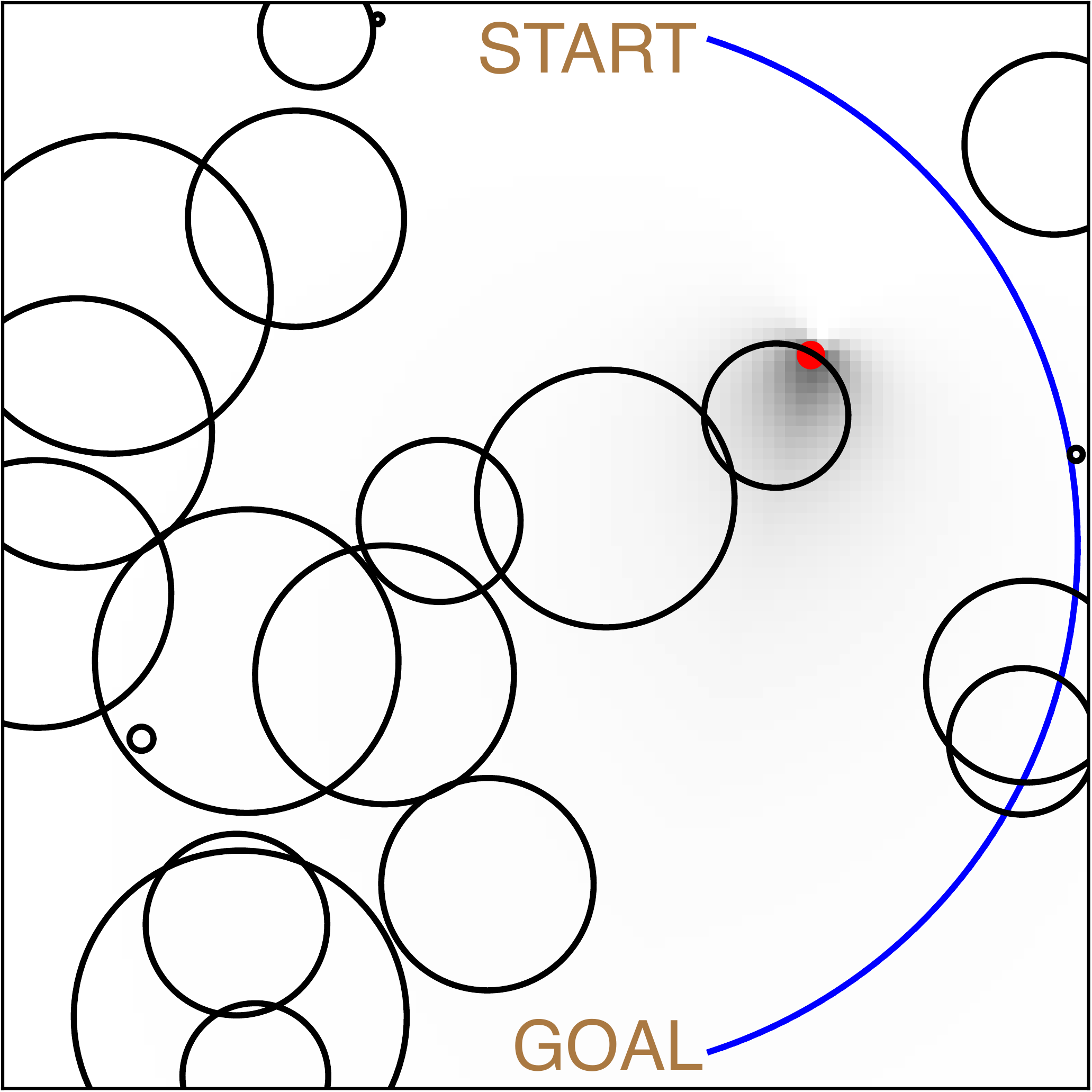} &
		\hspace{-1em}\includegraphics[width=0.117\textwidth]{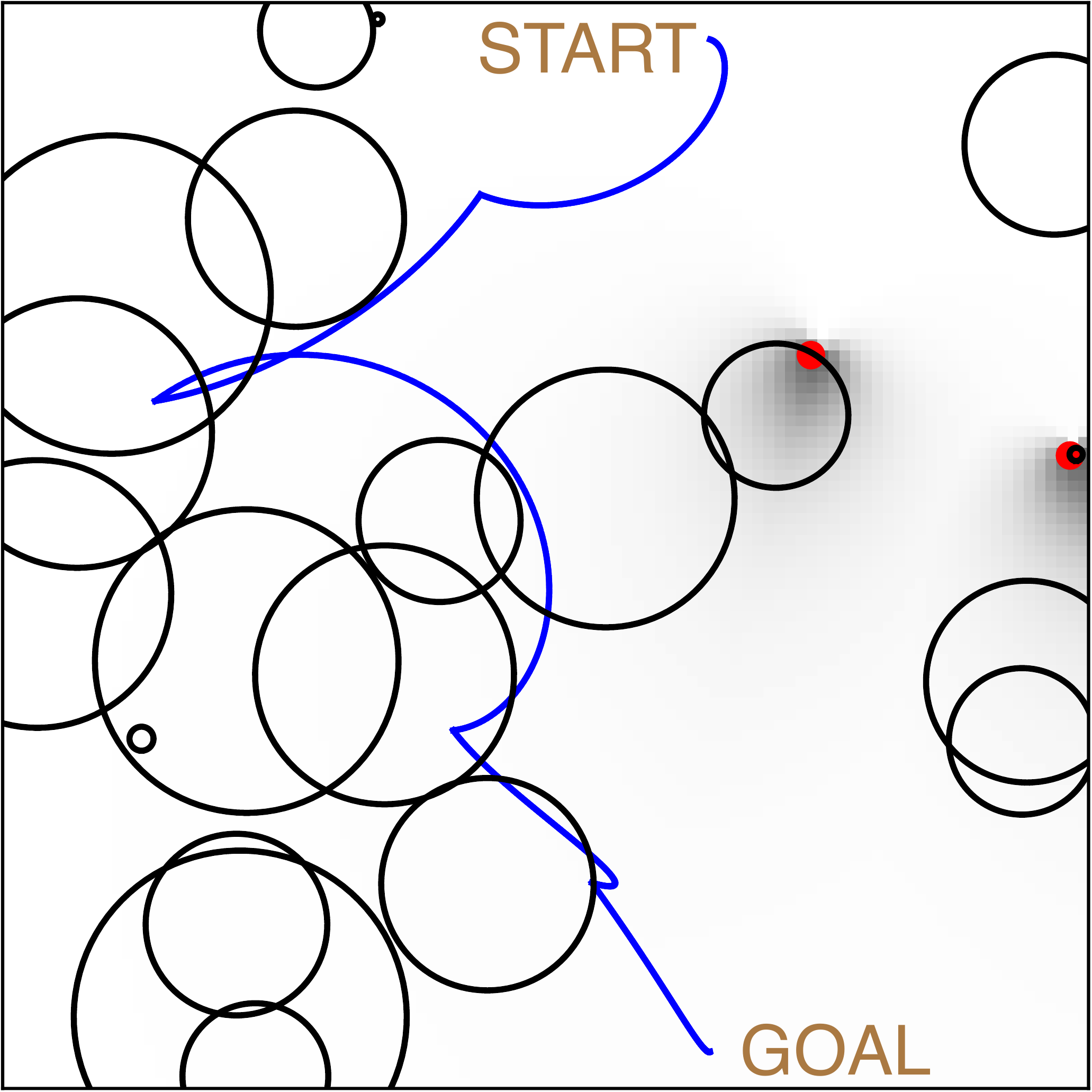}&
		\hspace{-1em}\includegraphics[width=0.117\textwidth]{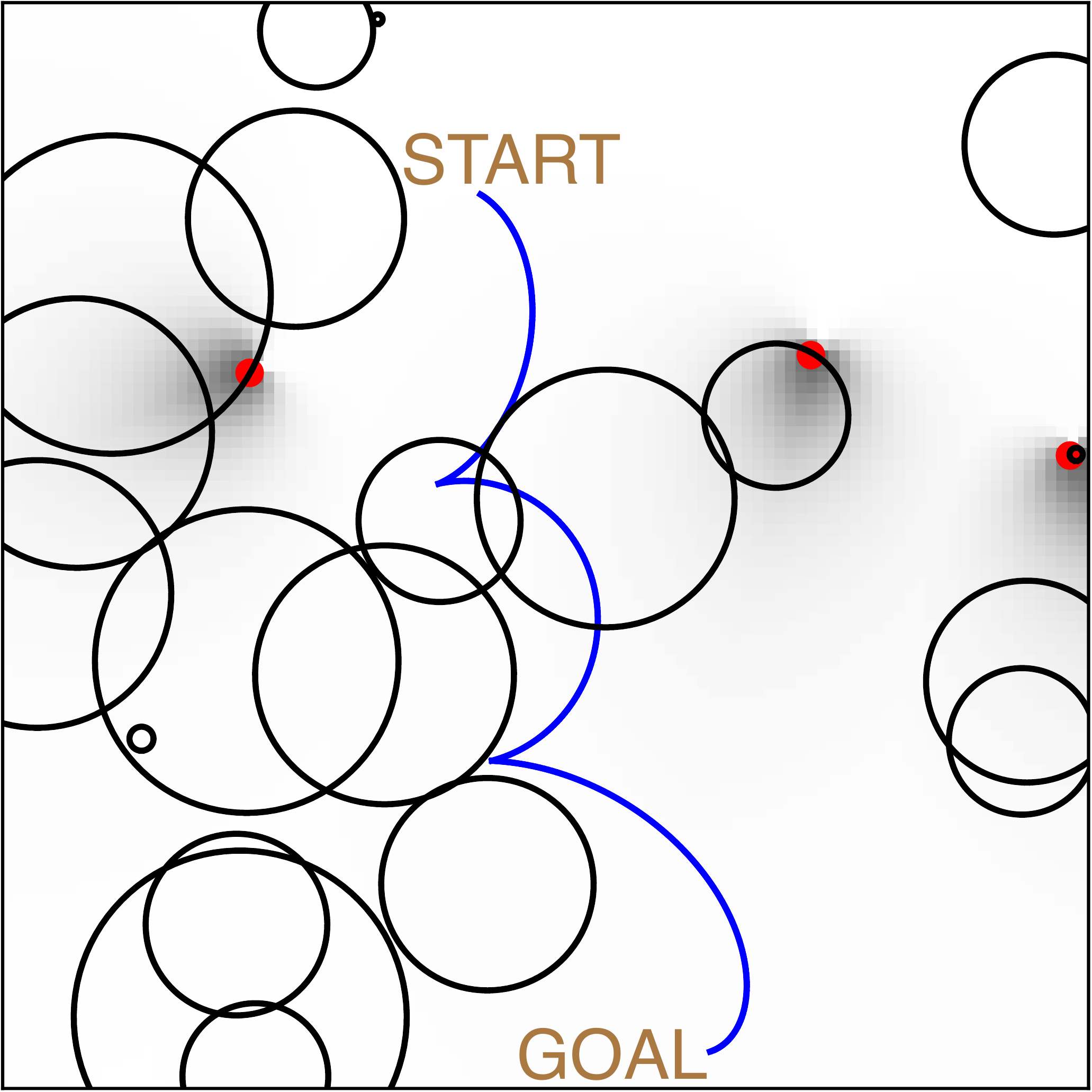}&
		\hspace{-1em}\includegraphics[width=0.117\textwidth]{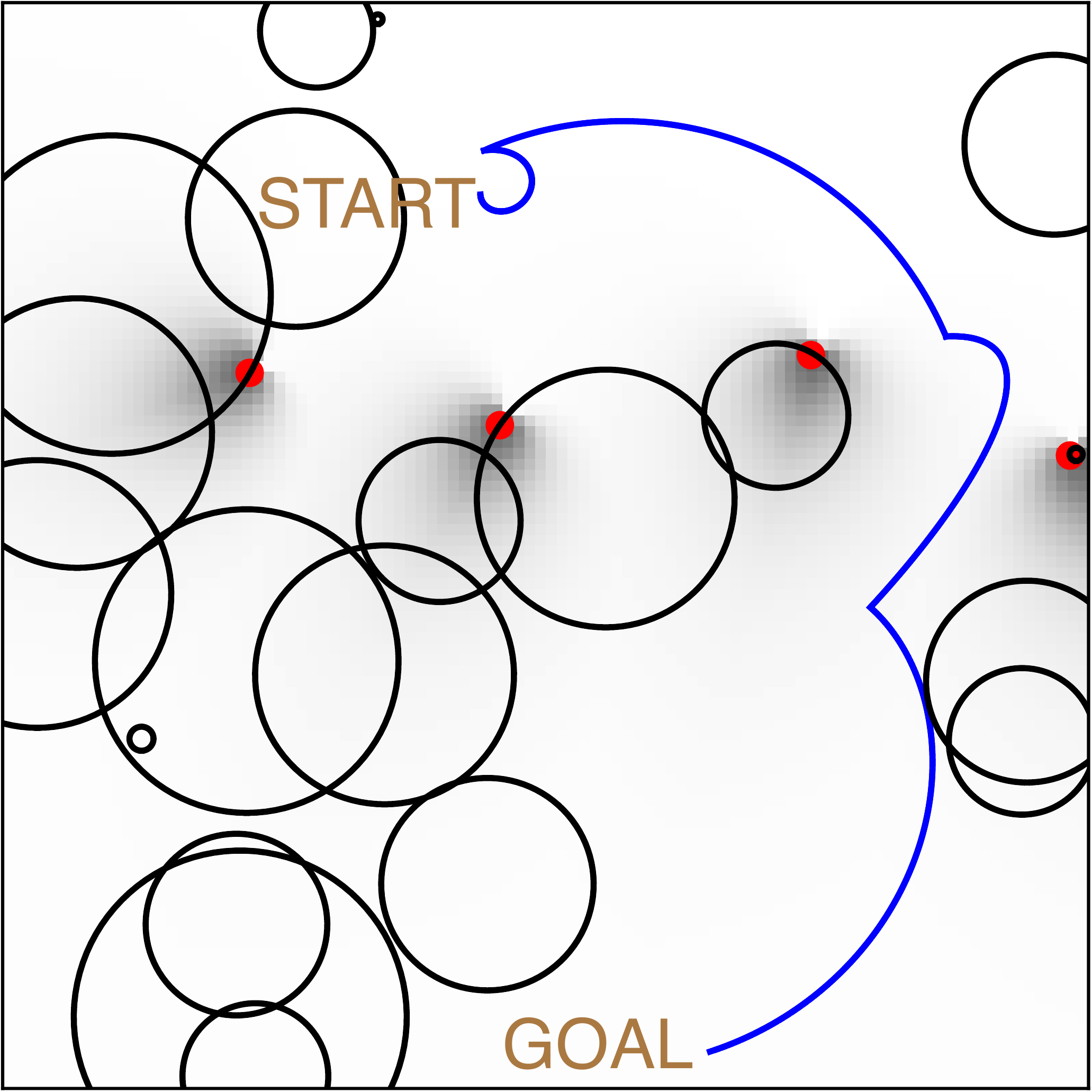}
	\end{tabular}
	\caption{Physical disentangling of an object. 
		\textbf{(Top)} A robot has grasped an object and needs to disentangle
		the object from other objects. In this case plastic and iron waste. The robot
		does not get sensory input except its own joint configuration. At each time step
		the robot plans a path in joint space, tries to execute the path, and collides with an obstacle.
		The collision is added to a list of collisions. We form a probability map from 
		the task space collisions that tells where the robot will likely collide with obstacles and
		use the map for choosing paths with low failure probability.
		\textbf{(Bottom)} Since visualizing the high-dimensional probability map for the real robot is challenging, 
		we show for illustration purposes 
		a sequence of path movements in 2-D task space simulation (we still plan
		in 7-DOF joint space). A planar multi-link robot
		uses our probabilistic disentangling approach to choose movements.
		Blue line denotes planned path, red dots previous collisions, and dark-light colors the probability map
		where dark color indicates high failure probability. Dark circles depict the real obstacles which are not visible to the robot.}
	\label{fig:overview}
\end{figure}

Robots have had a great impact on the manufacturing industry due to their
ability to autonomously handle large and heavy objects with high speed
and precision. For example, the car manufacturing industry is a
typical example for the benefits of robotic automation which is
nowadays a key component in the production pipeline of almost every
major car manufacturer. Robots have similarly large potential for
performing less structured tasks such as sorting and segregating heaps
of waste. However, applying robots to such domains is more
challenging, because it is no longer sufficient to execute
preprogrammed motions due to the uniqueness of the encountered
situations. In order to segregate waste, robots need to be able to
disentangle objects with unknown shapes. The objects may be entangled in
complex ways that are not fully perceivable due to limited sensory
information. Furthermore, the effects of disruptive actions, such as
pulling a given part of the heap, can not be accurately predicted due
to complex dynamical interactions and non-rigid objects.

Albeit challenging, applying robots for waste segregation can be
highly rewarding. For example, decommissioning nuclear waste which
is inherently dangerous for humans is an ideal task for robots. The Sellafield~(UK) nuclear site
is the largest nuclear site in Europe, containing 140 tonnes of civil
plutonium~\cite{wade2015sellafield} and 90,000 tonnes of radioactive
graphite~\cite{pearce2015shocking} and the cost of decommissioning was
estimated at 47.9 billions GBP in
2015~\cite{nao2015progress}. Decommissioning includes sorting and
segregating 69,600~m$^3$ of mixed intermediate waste, such that waste
with high level of radiation can be safely stored in expensive,
well-shielded containers, while avoiding unnecessary filling such
containers with waste with low level of radiation.

Targeted at this application, we propose a method for autonomous
disentangling of waste that is capable to deal with complex, entangled
objects based on limited sensory information and that does not require
models of the objects or of the dynamical interactions. We assume that the
end-effector of the robot has grasped a part of the heap and our goal
is to move the end-effector together with the grasped object away from
the heap. When such movement succeeds, we consider the object to be
disentangled from the heap. 

As fully planning such movements a priori is not feasible due to limited sensor data, we
adopt an adaptive planning approach that re-plans new movement
trajectories such that it avoids motions that are similar to motions
that failed in the past due to object entanglement. We therefore learn
a probability map over joint configurations that can be used to assess
the probability of failing along a given trajectory and use these
probabilities as a cost for a variant of the Rapidly exploring Random
Tree (RRT~\cite{lavalle01}) trajectory planner. Our main contribution is an approach that can remove an unknown object
from a set of unknown objects when the object configuration prevents a
simple movement separating the object from the other objects. We call
the process of removing an ``entangled'' object from other objects
``disentangling''.

Our approach does not need any sensory input which is crucial under
heavy occlusion, for example, in waste segregation. Moreover, in the
nuclear waste scenario, accurate tactile or vision sensors can be hard
to use due to potentially high nuclear radiation. Instead of requiring
informative sensors, object models, or actually seeing through
objects, our approach needs only information about the joint positions
of the robot to detect when movement is blocked. In practice, a
movement can be blocked either due to the robot being configured to
limit the amount of force, as in our experiments, e.g.\ using
impedance control, or a robot physically not being able to exert
sufficient force to move through an obstacle.

Our assumptions on the objects is minimal. Due to using a
probabilistic formulation of the disentangling task we do not have to
assume rigid objects, required by hard constraints, but we instead
assign success probabilities to movements and can cope with a
(moderately) changing environment: due to never having deterministic
probabilities the system has the chance to self-correct its model of
the world based on new observations. To summarize:
\begin{itemize}
\item We propose a new approach for disentangling objects based on a
  novel incremental probabilistic path planning formulation.
\item Our approach improves the robots knowledge of the environment
  incrementally until succeeding.
\item Due to the probabilistic formulation the approach works in an
  environment with both rigid and non-rigid unknown objects.
\item The approach does not need any sensors but relies solely on
  joint position information.
\end{itemize}

\section{Related Work}

In the task of disentangling an unknown object the robot needs to find
a motion or path that transports the entangled object out of the
entanglement. Path planning in a known environment is a widely studied
problem~\cite{kavraki1996probabilistic,lavalle01,lavalle2006planning,karaman06}
especially in robot motion planning~\cite{yang2010elastic} and path
planning of mobile robots~\cite{ge2000new,hu2004knowledge}. Path planning algorithms are
often based on the Rapidly explorating Random Tree
(RRT)~\cite{lavalle01} or the Probabilistic Roadmap
(PRM)~\cite{kavraki1996probabilistic}. RRTs iteratively build a tree of joint configurations where the root node corresponds to the start configuration and edges indicate that a direct movement between two configurations is feasible. At each iteration a new joint configuration is sampled and its closest joint configuration within the tree is determined. If a direct connection to the nearest configuration is feasible, the new configuration is added as its child. 
Once a node inside the tree is sufficiently close to the goal, a path from the start configuration to the goal can be recovered in reverse order by iteratively adding the parent node as via point.
PRMs build a graph by sampling random nodes and adding to the
graph and then finally finding the shortest path using a graph search
method such as Dijkstra's. Due to the requirements of planning in continuous space in reasonable computation time we adapt an RRT based planning approach but
the planner could be exchanged for another.

While the problem of robot manipulator motion planning has been
studied, motion planning in unknown environments with unknown obstacles is only little explored, especially for larger than two- or three-dimensional spaces. \cite{lumelsky87} provides an approach for path planning
with a point robot moving in 2D space containing unknown
objects. However, the approach in~\cite{lumelsky87} is inherently
limited to a 2-D space where the robot can travel around an obstacle
and is able to utilize sensory feedback to keep the robot close to the
obstacle.

The D*-path planning algorithm~\cite{stentz94} which takes inspiration
from the A* planning method~\cite{hart1968formal} can adapt to changes in the environment
and finds an optimal solution in the limit. However, D*-path planning is
limited in practice to discrete state spaces. \cite{van06} presents a PRM based approach for path
planning and replanning in dynamic environments where the agent
repairs its trajectory when new information is received.

Path planning in unknown and dynamic environments has been studied in
the context of mobile robot path               
planning~\cite{hoy15,tobaruela17}. \cite{wang05} provides a fuzzy
logic based approach for mobile robot path planning and demonstrates
the approach in path planning for mobile robots in a 2-D
environment. \cite{lei06} demonstrates a genetic algorithm based
approach to path planning for mobile robots in a dynamic unknown
environment. \cite{sfeir11} utilizes artificial potential fields
(APFs)~\cite{hwang1992potential} where obstacles repel and goals
attract the robot. \cite{montiel15} uses the Bacterial Evolutionary
Algorithm (BEA) together with APFs for mobile robot navigation.

In human-robot interaction, the robot often needs to be able to avoid
collisions with a hard to predict human. Ideally the robot would not
need to make prior assumptions about the human's location but could
learn this online. However, due to the safety requirements human-robot
interaction requires in practice (stochastic) models or predictions of
human behavior \cite{mainprice2013human}. In physical object
disentangling the robot does not usually need to take safety similarly into
account and thus prior models are not crucial.

\textbf{Object disassembly.} Disentangling an object away from objects
with which it is entangled corresponds to disassembly of an object
from a structure. \cite{cortes2008disassembly} perform path planning
for disassembly of complex objects. However, to the best of our
knowledge disassembly of unknown objects in an unknown environment has
not been performed before. In this paper, our experimental setup
focuses on disentangling waste objects but in future work our approach
could be applied to disassembly of complex unknown structures.

\textbf{Waste segregation.} Waste segregation is one of the
applications motivating our work on object
disentangling. State-of-the-art waste segregation relies on
classifying and picking up waste that is not
entangled~\cite{lukka2014zenrobotics,kujala2016classifying,paulraj2016automated,gundupalli2017review}. Being
able to disentangle different types of waste would allow more
efficient reuse of materials. Moreover, being able to disentangle
hazardous materials such as nuclear waste~\cite{nao2015progress} can
be crucial for safe decommissioning.

\section{Approach}

In this paper, we do not take grasping into account but assume
that an object has been already grasped. Grasping is a widely studied
problem and there is a multitude of approaches for grasping~\cite{saxena08,maitin10,lenz15}. Our goal is to disentangle
the object. The main \textbf{research question} we investigate is: How
to disentangle an unknown object which is firmly planted in the
gripper from other unknown objects? To answer this research question we
start by considering the properties of the problem and then discuss an
approach that takes advantage of these properties.

\textbf{Problem characteristics.} We focus on general disentangling
problems where we have both non-rigid and rigid unknown objects. Our
approach can be applied to, for example, waste segregation with various
objects. The goal is to move the robot's joints into a goal
configuration. The robot always knows its current joint configuration 
and can thus compare its current configuration to the desired one.
This way the robot is able
to detect when a movement is blocked but we do not assume any other
sensory input.
This kind of decision making problem can be formalized
as a partially observable Markov decision process
(POMDP)~\cite{astrom65,kaelbling98} since we have a sequential
decision making problem under object and environment uncertainty and
only get partial observations about the environment state. In particular,
the robot does not observe in which directions it is able to move.

\textbf{Assumptions.} We assume a moderately stationary environment:
objects do not move. Since we do not use hard constraints the robot
may always plan a path to the goal and may succeed even in a
non-stationary environment. However, we do not take non-stationarity
explicitly into account.

\textbf{Proposed approach.} Instead of doing a full POMDP solution
which is computationally intractable~\cite{papadimitriou87}, at each
decision step the robot greedily tries to find a path to the goal
position based on the previously obtained information. Since we get new
information with each movement our model of the environment gradually
improves and we have a better chance of succeeding. Moreover, since
we obtain information about blocked movements, the robot will learn to
circumvent blockades and thus optimal information gathering, provided
by a full POMDP solution, is not required in practice. The robot
remembers every movement and stores information about blocked movements. The robot creates a
probability map from the failed (blocked) movements that assigns to any given position an assumed probability of being in collision. 
In principle, the probability map can be computed directly based on the joint configurations. However, for our robot experiments we assume that collisions are always caused by the object entanglement and not by the links of the robot. We thus compute the probability map in end-effector space based on the forward kinematics.
We assume the probability of collision to be high, if the end-effector pose is similar to the pose of a previously blocked movement. However, we do not only consider the distance between the given pose and the colliding poses, but also take into account the movement direction during the previous collisions. Intuitively, the probability of colliding should increase if the end-effector position lies along the direction of the colliding movement.
We next discuss the proposed approach in more detail.

\subsection{Details}
Algorithm~\ref{alg:probdis} shows our proposed approach to object
disentangling in pseudo-code format (further down we provide symbol definitions).
The task is to find a path from the current configuration $c_{\textrm{CURRENT}}$ to a
goal configuration $c_{\textrm{GOAL}}$ where the object is fully disentangled.
At each time step the robot plans
a path based on the probability map $\mathcal{M}$ and executes the
path. In case the movement arrives at the goal disentangling ends. If
the movement fails, the robot moves back to the last via point where
movement succeeded.
In the experiments with a compliant (limited force) robot arm we
assume failure if the robot ends at joint configuration $c_{\textrm{CURRENT}}$, above a certain distance $d_{\textrm{DIST}}$ from
the desired joint configuration $c_{\textrm{DESIRED}}$. That is,
if the sum of elements in $|c_{\textrm{DESIRED}} - c_{\textrm{CURRENT}}|$ is greater
than $d_{\textrm{DIST}}$.

The robot adds the failure (blocked movement) to the list of movements
which defines $\mathcal{M}$ and starts replanning from the actual
position of the robot.
Note that we call $\mathcal{M}$ a probability map since the
previous movements are used to compute pseudo-probabilities which can
be visualized as a map.
For simplicity the path planning consists of
a bidirectional version of RRT~\cite{lavalle01} based on
RRT-connect~\cite{kuffner2000rrt} and 
includes rewiring from RRT*~\cite{karaman06,klemm2015rrt} but can
be easily exchanged or extended with other path planners that allow
arbitrary cost functions. The cost function comes from our
probabilistic formulation which we discuss further down in more
detail.

\SetKwProg{Fn}{}{}{}
\begin{algorithm}[ht!]
        \caption{Probabilistic Disentangling (ProbDis). At each time step 
		use probabilistic RRT to find path with low failure probability and
		execute the path. Add resulting collision to the list of collisions which 
		defines the failure probability map $\mathcal{M}$
		which is used for RRT planning.}
	\label{alg:probdis}
	\SetKwFunction{ProbDis}{ProbDis}
	\SetKwFunction{ProbRRT}{ProbRRT}
	\SetKwFunction{RobotMotion}{RobotMotion}
	\SetKwFunction{Connect}{Connect}
	\SetKwFunction{Cost}{Cost}
	\Fn{\ProbDis{}}{
		$\mathcal{M} = \varnothing$\\
		\While{{Entangled, below time limit, $|\mathcal{M}| < N_{\textrm{ITER}}$}}{
			// Create RRT tree from start configuration\\
                        $G_{\textrm{START}} =$ \ProbRRT($M, c_{\textrm{START}}, c_{\textrm{GOAL}}$)\\
			// Create RRT tree from goal configuration\\
                        $G_{\textrm{END}} =$ \ProbRRT($M, c_{\textrm{GOAL}}, c_{\textrm{START}}$)\\
                        // Connect the RRT trees\\
			$G =$ \Connect{$\mathcal{M}, G_{\textrm{START}}, G_{\textrm{END}}$}\\
			// Using $G$ find minimum cost start-to-goal path \\
			path = $\argmin_{\textrm{path}} $\Cost{$\mathcal{M}$, path}\\
                        // Move robot arm along path until failure\\
			failure = \RobotMotion(path)\\
			$\mathcal{M} = \mathcal{M}$ $\cup$ failure
		}
	}
\end{algorithm}
\begin{algorithm}[ht!]
	\caption{Probability map based RRT. Use map of movement
		probabilities $\mathcal{M}$ to find new nodes.}
	\label{alg:ProbRRT}
	\SetKwFunction{ProbRRT}{ProbRRT}
	\SetKwFunction{RandomConfiguration}{RandomConfiguration}
	\SetKwFunction{ClosestConfiguration}{ClosestConfiguration}
	\SetKwFunction{NewConfiguration}{NewConfiguration}
	\SetKwFunction{RewireNeighbours}{RewireNeighbours}
	\Fn{$G$ = \ProbRRT{$\mathcal{M}, c_{\textrm{START}}, c_{\textrm{GOAL}}$}}{
                $G = \varnothing$\\
                \For{$k = 1$ to $K$}{
			$c_{\textrm{R}} = $\RandomConfiguration($\mathcal{M}, G, c_{\textrm{GOAL}}$)\\
			$c_{\textrm{N}} = $\ClosestConfiguration($\mathcal{M}, G, c_{\textrm{R}}$)\\
                        // $c$ as a linear interpolation between $c_{\textrm{R}}$ and $c_{\textrm{N}}$\\
			$c = $\NewConfiguration($c_{\textrm{R}}$, $c_{\textrm{N}}$)\\
			$G = G \cup c$\\
			\RewireNeighbours($G, c$)\\
		}
		\Return $G$
	}
\end{algorithm}
\begin{algorithm}[ht!]
	\caption{Generate random configuration that is far away from previous configurations as
		measured by task space distance.}
	\label{alg:RandomConfiguration}
	\SetKwFunction{RandomConfiguration}{RandomConfiguration}
	\SetKwFunction{Distance}{Distance}
	\Fn{\RandomConfiguration{$\mathcal{M}, G, c_{\textrm{GOAL}}$}}{
		$w \sim Uniform(0,1)$\\ 
		\If{$w < P_{\textrm{GOAL}}$}{
			\Return $c_{\textrm{GOAL}}$\\
		}
		Sample a set of configurations $C_{\textrm{RAND}}$ uniformly\\
		// Chooce $c_{\textrm{RAND}}$ which is furthest away, in task\\
                // space, from previous RRT nodes $G$\\ 
		$c_{\textrm{RAND}} = \argmax_{c \in C_{\textrm{RAND}}} \Distance(G, c)$\\
		\Return{$c_{\textrm{RAND}}$}\\
	}
\end{algorithm}
\begin{algorithm}[ht!]
	\caption{Find closest configuration to $c$ from $G$ given probability map
		$\mathcal{M}$.}
	\label{alg:ClosestConfiguration}
	\SetKwFunction{ClosestConfiguration}{ClosestConfiguration}
	\SetKwFunction{Cost}{Cost}
	\Fn{\ClosestConfiguration{$\mathcal{M}, G, c$}}{
		${c_{\textrm{NEW}}} = \argmin_{c^*_{\textrm{NEW}} \in G} $
		\Cost($\mathcal{M}, c^*_{\textrm{NEW}}, c$)\\
		\Return ${c_{\textrm{NEW}}}$\\
	}
\end{algorithm}
Algorithm~\ref{alg:probdis} shows the main disentangling loop,
Algorithm \ref{alg:ProbRRT} creates an RRT tree and
Algorithms \ref{alg:RandomConfiguration}, \ref{alg:ClosestConfiguration},
and \ref{alg:Cost} perform the standard RRT functions of selecting a
random and closest configuration, and estimating the cost for a path.

The ``Connect'' procedure in Algorithm~\ref{alg:probdis} connects
each leaf node in $G_{\textrm{END}}$ to the node in
$G_{\textrm{START}}$ with minimum path failure probability. Path
failure probabilities are estimated using the probability map
$\mathcal{M}$ as detailed in Algorithm~\ref{alg:Cost}. Therefore,
``Connect'' results in worst case computational complexity of
$O(|G_{\textrm{END}}| |G_{\textrm{START}}| N_{\textrm{DISCRETIZE}} F_{\textrm{COMP}})$
where $N_{\textrm{DISCRETIZE}}$ is the maximum number of failure
probability computations in Algorithm~\ref{alg:Cost} and $F_{\textrm{COMP}}$
is the complexity of a failure computation.
\begin{algorithm}[ht!]
	\caption{Cost for moving from $c$ to $c_{\textrm{NEW}}$
		given probability map $\mathcal{M}$.}
	\label{alg:Cost}
	\SetKwFunction{Cost}{Cost}
	\SetKwFunction{ProbFail}{ProbFail}
	\Fn{\Cost{$\mathcal{M}, c_{\textrm{NEW}}, c$}}{
		$D = ||c - c_{\textrm{NEW}}||$\\
		// Discretize path using a discretization\\
		// distance of $\Delta$\\
		\For{$c_{\textrm{TEST}} \in [\hat{c} | c + \Delta \cdot k, k = 1 \dots D / \Delta]$}{
			// Compute failure probability for $c_{\textrm{TEST}}$\\
			// using Eq.~\ref{eq:prob_fail}\\
			$p^k_{\textrm{FAIL}} = $ \ProbFail{$\mathcal{M}, c_{\textrm{TEST}}$}\\
		}
		$p_{\textrm{FAIL}} = P(e = \textrm{FAIL} | p^{1 \dots K}_{\textrm{FAIL}})$\\
		\Return $p_{\textrm{FAIL}}$\\
	}
\end{algorithm}

\textbf{Definition of $x$, $c$, $\mathcal{M}$, a path, and movement failures.}
We will now discuss how
movement failures are represented, how we compute
pseudo failure probabilities and how we estimate path costs for path
planning. Note that in the robot experiments the robot moves in joint
configuration space but we handle failures in task space.
That is, when computing failure probabilities we transform joint
configurations to task space using forward kinematics.
$x$ refers always to task space coordinates. For 2-D coordinates
$x \in \mathbb{R}^2$, for 3-D $x \in \mathbb{R}^3$, and for 3-D with
orientation $x \in \mathbb{R}^6$.
$c \in \mathbb{C}$ refers to joint
configurations where $\mathbb{C}$ is the joint configuration space.
A single move $m = (x_{\textrm{START}},
x_{\textrm{END}})$ consists of the start point $x_{\textrm{START}}$
and end point $x_{\textrm{END}}$. A path $(m_1, \dots,
m_{N_{\textrm{PATH}}})$ is a sequence of $N_{\textrm{PATH}}$ moves. We
define $v_{\textrm{FAIL}}$ as the vector representing the direction
the robot was moving towards when it was blocked. Moreover, we define
a failure $f = (x_{\textrm{FAIL}}, v_{\textrm{FAIL}})$ using the point
$x_{\textrm{FAIL}}$ where the robot was blocked together with the
vector $v_{\textrm{FAIL}}$. $f$ is stored in task space
coordinates. 
Every executed robot path results in a set of successful
movements and a single failure in case the robot did not reach the
goal. We define $\mathcal{M}$ as a set of $N_{\textrm{FAIL}}$
failures: $\mathcal{M} = (f_1, \dots, f_{N_{\textrm{FAIL}}})$.

\textbf{Failure probabilities, cost function.} We heuristically
estimate the probability $P(e = \textrm{FAIL} | c, F)$ of a failure
event $e = \textrm{FAIL}$ ($e$ denotes event, $\textrm{FAIL}$ failure)
of a specific configuration $c$ given previous 
failures $F$.
We do not assume knowledge of how failure probabilities could be
interpolated from several failures. Instead, we estimate the probability of failure for a given
set $F$ as the maximum probability of failure over all individual
failures in $F$:\\
$P(e = \textrm{FAIL} | c, F) = \max_{f \in F} P(e = \textrm{FAIL} | c, f).$
For simplicity, we assume that the probability of failure given a previous failure $f$ mainly depends on the end-effector position but can be halved in the best case for fully dissimilar end-effector orientations, that is,
\begin{align}
P(&e = \textrm{FAIL}| c, f) = \nonumber\\
&P_{\textrm{COORD}}(e = \textrm{FAIL}| c, f) (1 - 0.5 D_{\textrm{ORIENT}}(c,f)),%
\label{eq:prob_fail}
\end{align}
where $P_{\textrm{COORD}}$ estimates the probability of failure using only the given end-effector position, but not its orientation, and $D_{\textrm{ORIENT}}(c,f)$ measures the distance between the given end-effector orientation and the end-effector orientation during the failure $f$, normalized to the range $[0,1]$.

We compare the end-effector orientations based on the geodesic distance between
two quaternions defined in~\cite{huynh2009metrics},
\begin{equation}
D_{\textrm{ORIENT}}(c,f) = \frac{2}{\pi} \arccos(|x_{\textrm{ORIENT}}^T \cdot x_{\textrm{ORIENT}, \textrm{FAIL}}|),
\label{eq:D_orient}
\end{equation}
where we introduced the term $\tfrac{2}{\pi}$ for normalization.

We model $P_{\textrm{COORD}}(e = \textrm{FAIL}| c, f)$ based on the squared distance between the given end-effector position  and the end-effector position during failure, $D_{\textrm{FAIL}} = ||x_{\textrm{COORD}} - x_{\textrm{COORD}, \textrm{FAIL}}||^2$. Furthermore, we want to take into account that the probability of failure is usually much larger in the direction of movement $v_{\textrm{FAIL}}$. Hence, we compute the angle $\alpha$ between the failed movement direction $v_{\textrm{FAIL}}$ and the direction from current Euclidean coordinate to the failed Euclidean coordinate $x_{\textrm{COORD}} - x_{\textrm{COORD}, \textrm{FAIL}}$. This angle is at its maximum, if the current coordinate $x_{\textrm{COORD}}$ was in front of the failed Euclidean coordinate $x_{\textrm{COORD}, \textrm{FAIL}}$ and its minimum if it was behind.

\begin{figure}[thbp]
  \centering
  \hspace{-1em}\includegraphics[width=0.25\textwidth]{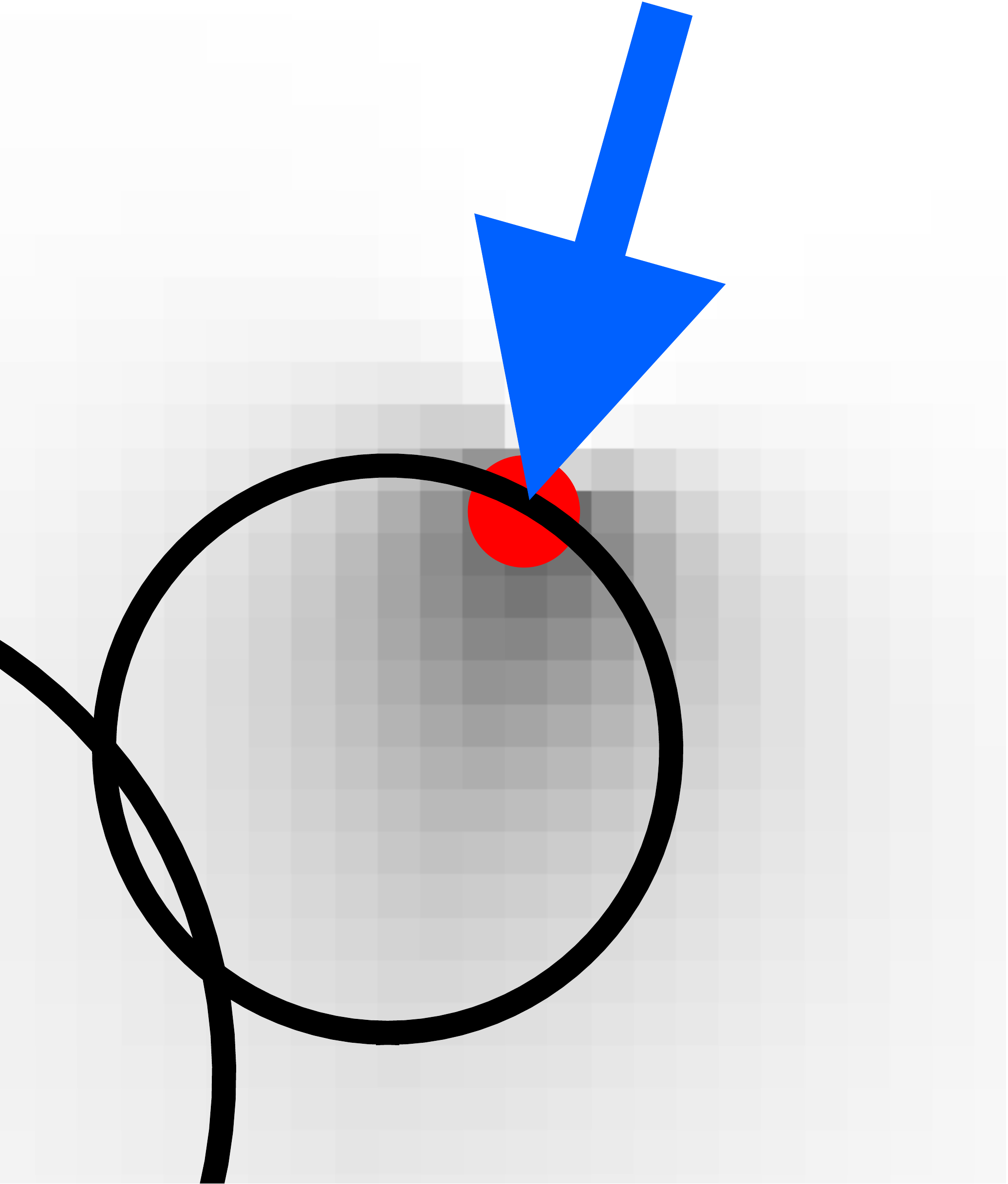}
  \caption{2-D example of how part of the failure probability map
    looks after a collision. Here, the robot collided at the red dot with the unseen
    circle obstacle when moving in the direction of the
    blue arrow. The failure probability behind the collision is high,
    indicated by dark color, and in front of the collision low as
    indicated by a light color.}
  \label{fig:probability}
\end{figure}

Based on preliminary experiments, we found the following model to produce sensible results,
\begin{equation}
P_{\textrm{COORD}}(e = \textrm{FAIL}| c, f) = |\alpha|^3 / \pi^3 \cdot 1 / (1 +
D_{\textrm{FAIL}} C_{\textrm{FAIL}}),
\end{equation}
where $C_{\textrm{FAIL}}$ is a constant. Note that
$P_{\textrm{COORD}}(e = \textrm{FAIL}| c, f)$ is always between zero
and one.
Fig.~\ref{fig:probability} shows that our model qualitatively 
makes sense.
For estimating the probability of failure for a path, assuming independence of failures
along the path, we would like to
compute the geometric product integral over the failure probability density of the path.
To approximate this, we discretize the path into $K$ segments
and transform the joint configuration at each point along the discretized path into
task space using forward kinematics. We then estimate path failure
probability $p^{k}_{\textrm{FAIL}}$ for each point $k$ using Eq.~\ref{eq:prob_fail} defined above and
distance between consecutive points along the path (corresponds to rectangle rule in numerical product integration).
We then estimate path failure
$P(e = \textrm{FAIL} | p^{1 \dots K}_{\textrm{FAIL}})$:
\begin{equation}
	P(e = \textrm{FAIL} | p^{1 \dots K}_{\textrm{FAIL}}) = \prod_{k=1}^K p^{k}_{\textrm{FAIL}}
	= \exp \sum_{k=1}^K \log p^{k}_{\textrm{FAIL}},
\end{equation}
where transforming the product into a sum of log terms can be important as part of a practical implementation
to prevent underflow of floating point numbers.

\section{Experiments}

We performed experiments both in simulated randomly constructed 2-D
and 3-D environments and with a 7-DOF KUKA LBR iiwa R820 robot arm. With the
robot arm we disentangle an object from real plastic and iron waste
and move the object out of a simple cardboard maze. We also
disentangle a fluffy toy-bunny which is jammed in a cardboard box.
In each run, we allow the robot to execute at most
$20$ paths and measure the success rate.
Success means arrival at the goal configuration $c_{\textrm{GOAL}}$.
In the experiments, we have $K = 100$, $\Delta = 0.04$,
$N_{\textrm{ITER}} = 20$, $|C_{\textrm{RAND}}| = 100$, $P_{\textrm{GOAL}} = 0.1$,
$d_{\textrm{DIST}} = 0.04$, and no time limit.

\subsection{Simulation}

\begin{figure*}[thbp]
	\centering \vspace{0.5em}
	\begin{tabular}{cccc}
		\hspace{-0.5em}\includegraphics[width=0.245\textwidth]{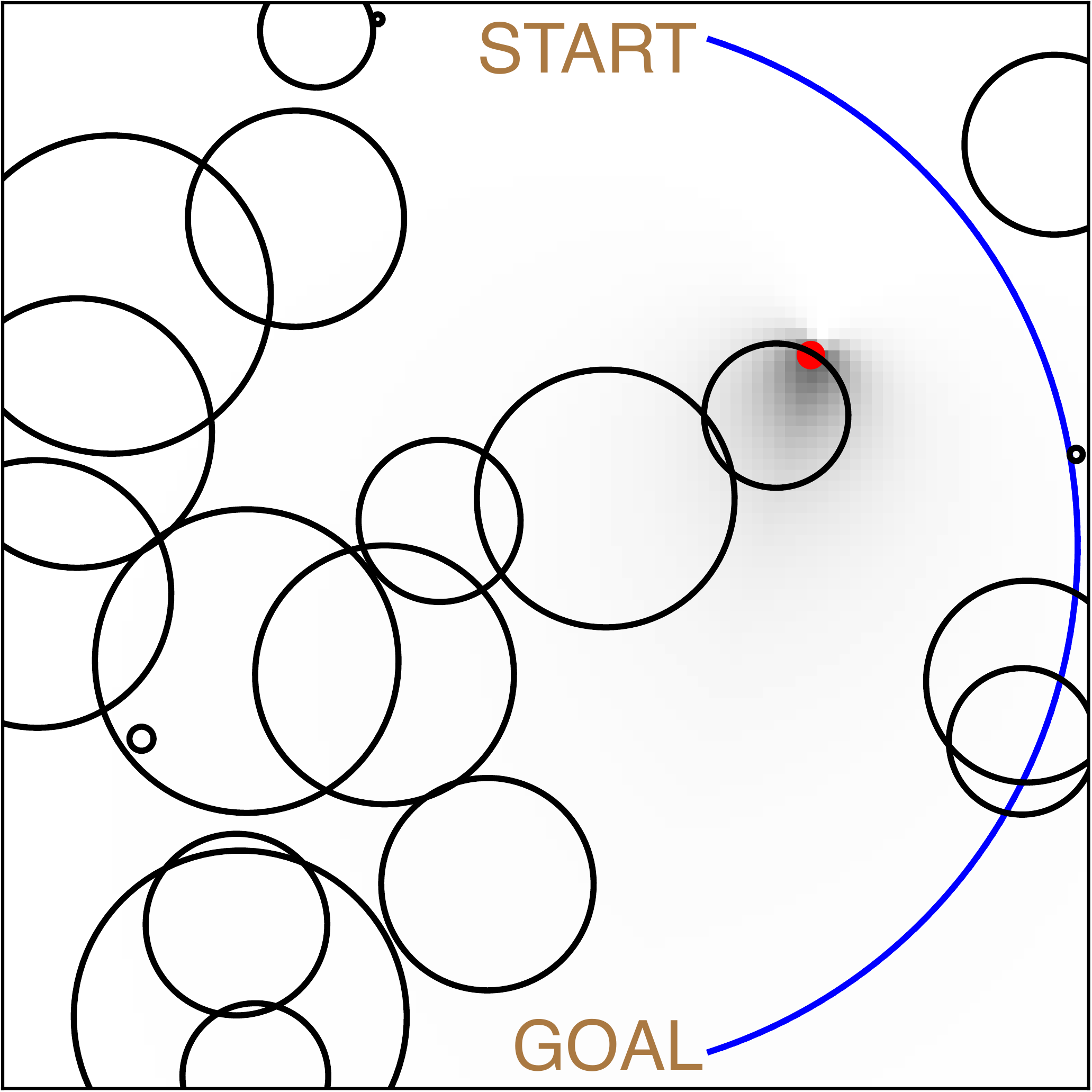} &
		\hspace{-1em}\includegraphics[width=0.245\textwidth]{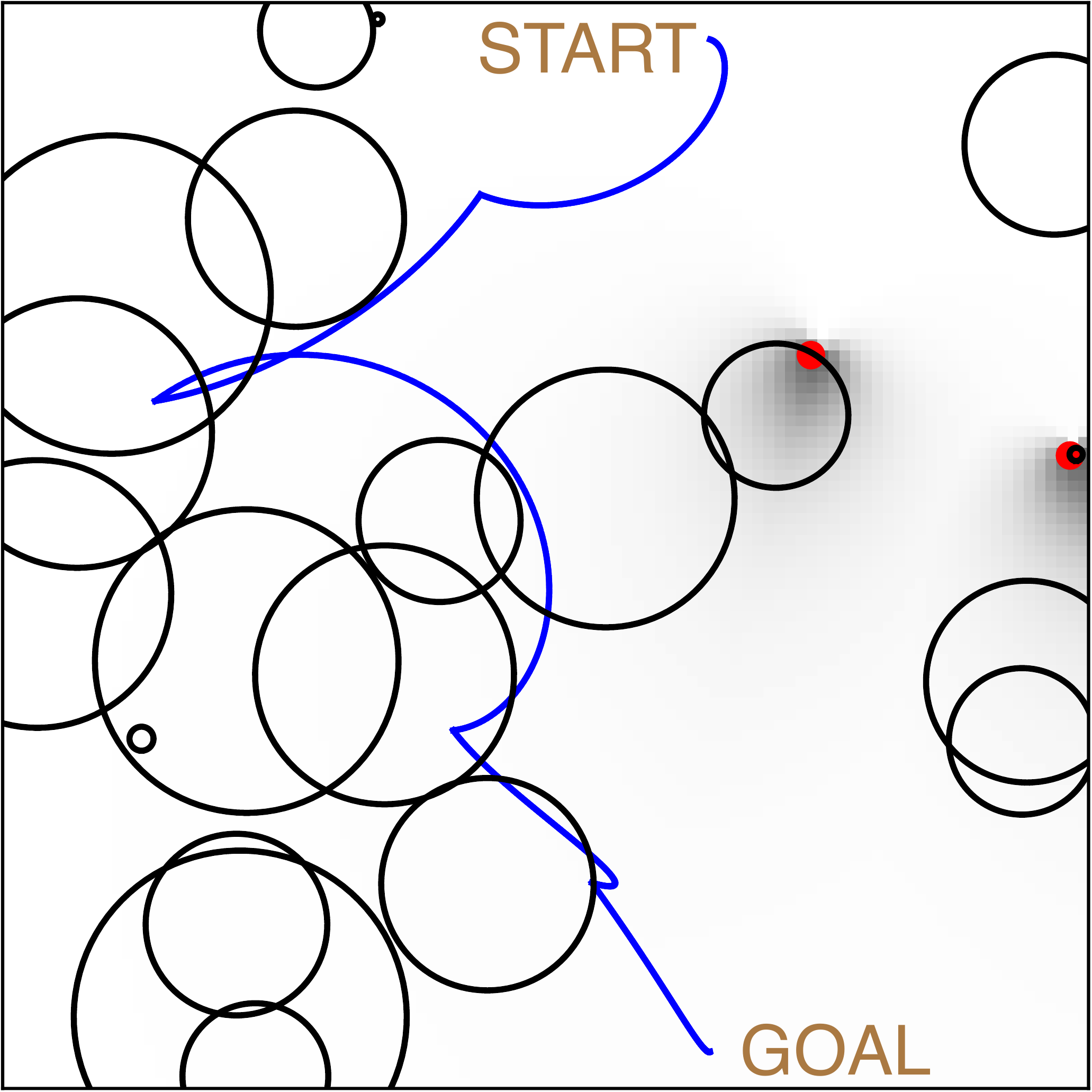} &
		\hspace{-1em}\includegraphics[width=0.245\textwidth]{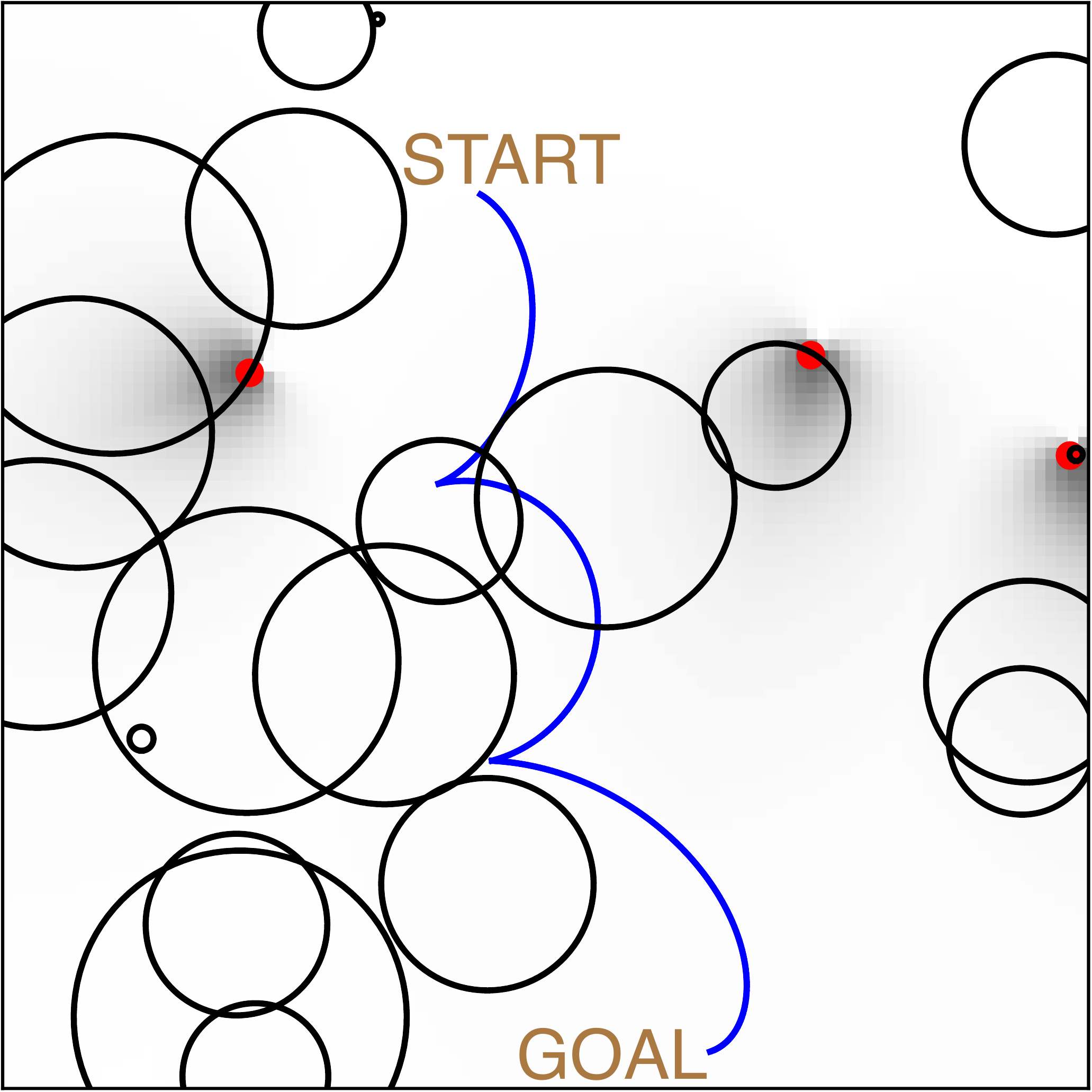} &
		\hspace{-1em}\includegraphics[width=0.245\textwidth]{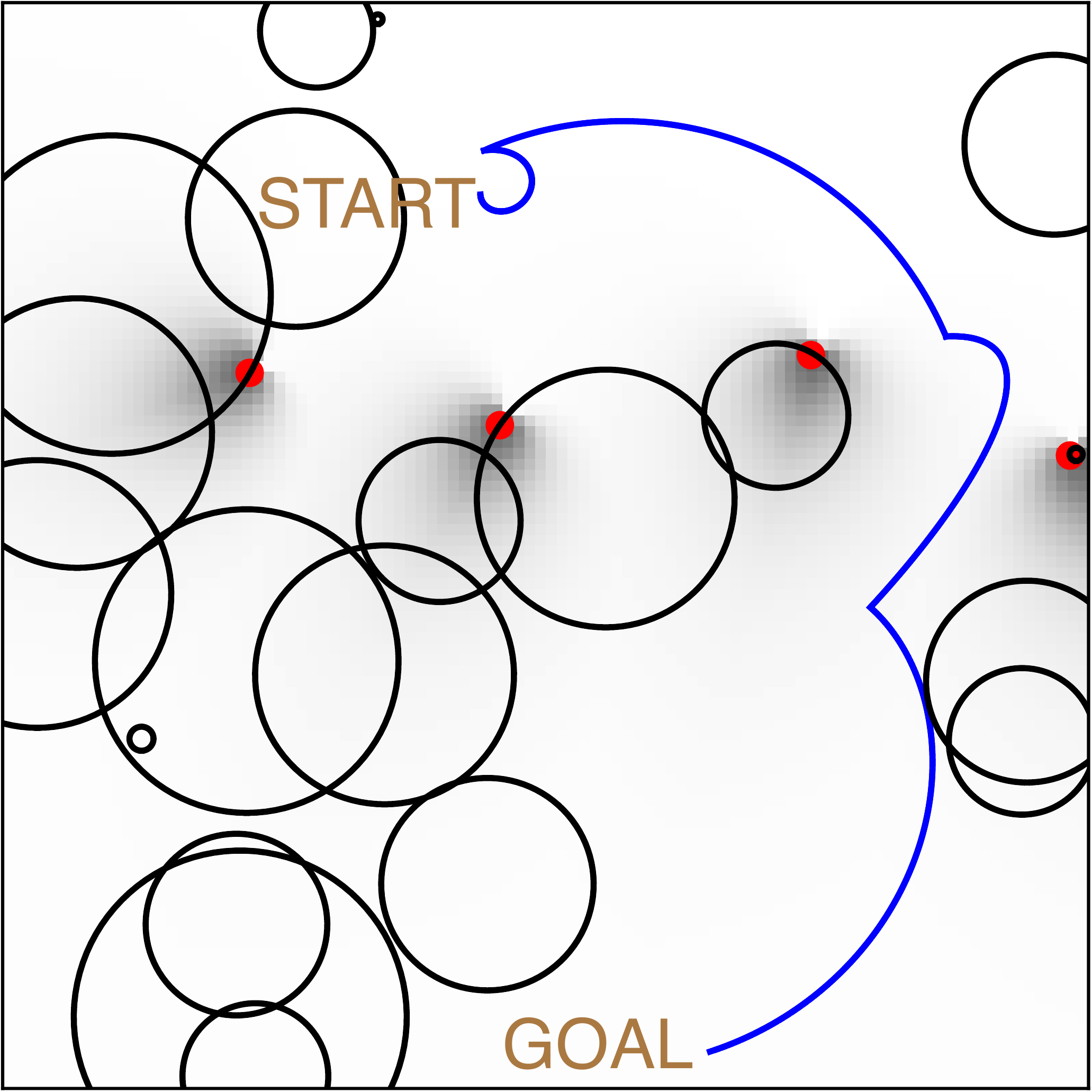}
	\end{tabular}
	\caption{Visualization of a sequence of disentangling movements of a planar multi-link robot in 2-D simulation.
		At each time step the robot
		plans a path in joint space, shown by blue color in task space, to the goal configuration and tries to execute the path. 
		Red dots show where
		the robot has previously collided with obstacles in task space.
		The robot does not observe the ``real obstacles'' shown as black circles but it can sense its current joint configuration.
		The robot's probability map $\mathcal{M}$ for collisions is shown with dark-light colors. Dark color represents
		high collision probability and white color low collision probability. In this image sequence, the
		robot hits obstacles four times before finally arriving at the goal in the last image. As can be seen, the robot always
	tries to circumvent areas with high collision probability.}
	\label{fig:simulation_visualization_2D}
\end{figure*}
\begin{figure}[thbp]
	\centering
	\includegraphics[width=0.48\textwidth]{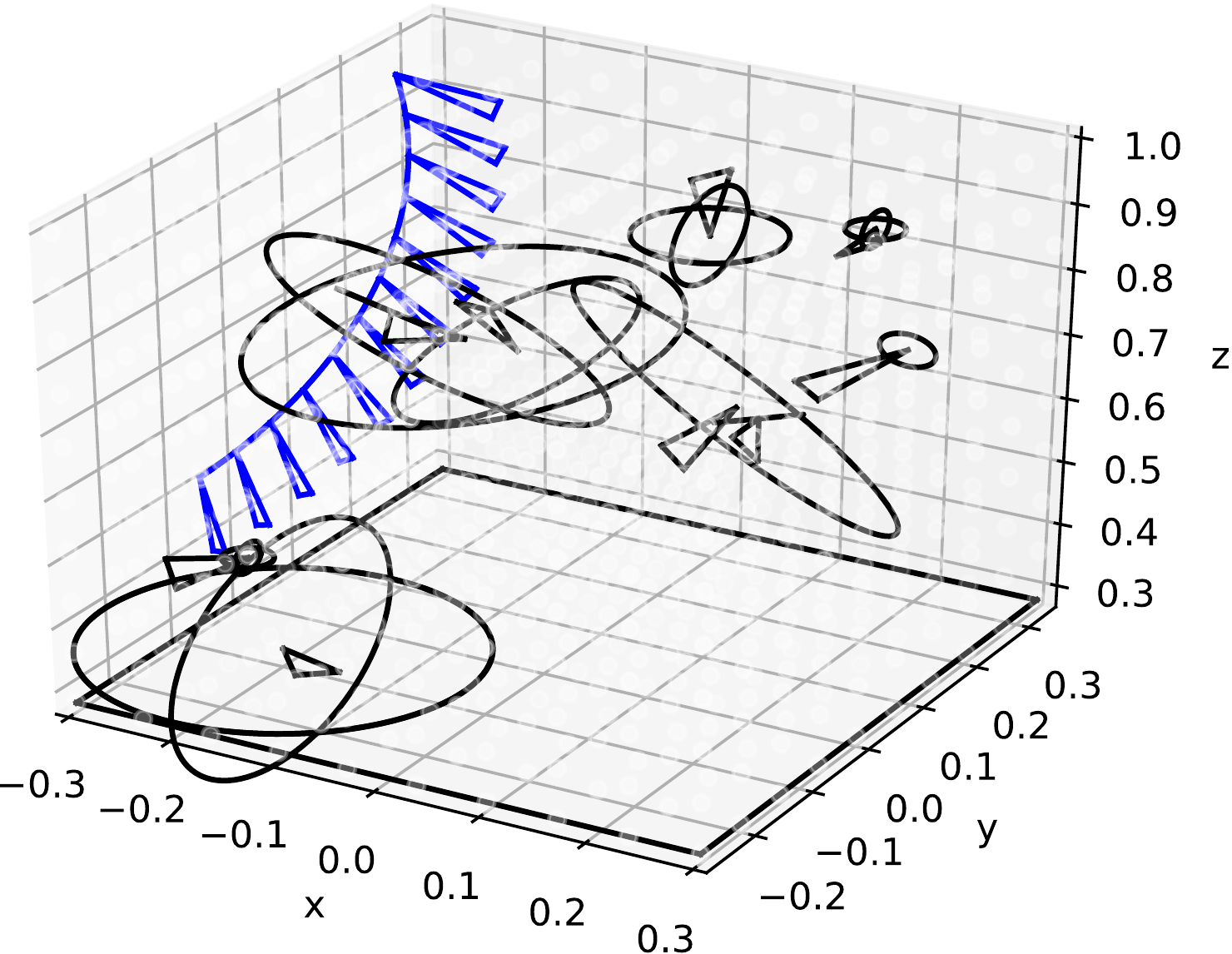}
        \caption{Visualization of 3-D disentangling simulation. Blue
		 line depicts planned end-effector
		 trajectory. Triangles show orientation of
		 end-effector and obstacles. The robot does not
		 observe the obstacles, only its joint configuration.
		 Obstacles are either discs, depicted by circles, or
		 balls, depicted by two orthogonal circles. Obstacle
		 size varies.
                 Obstacles can be circumvented by using
		 an end-effector orientation dissimilar from the
		 obstacle orientation.
                 Making orientation in this way important is
		 motivated by flexible objects which may allow you to
		 pull the object and slide along the object with a
		 suitable orientation.  Note also that we evaluate the
		 approaches with and without obstacle orientation.}
	\label{fig:simulation_visualization_3D}
\end{figure}
In the experiments, the proposed algorithm always plans in 7 dimensional joint configuration space
and checks for collisions in task space. In the simulations, we run experiments
(1) in two dimensional task space where the end-effector is just a point mass
at the end of a planar 7-DOF multi-link robot arm, (2) with a 3-D point mass at the
end of a KUKA LBR 7-DOF robot arm, and (3) where the 3-D task space also includes
the orientation of the end-effector. Case (1) is beneficial for visualizing the proposed
approach. The real robot experiments use case (3).
When checking for collisions or collision probability, in cases (1) and (2), the orientation of the end-effector is not taken into account, that is, $D_{\textrm{ORIENT}}(c,f)$ in Eq.~\ref{eq:D_orient} becomes zero.
Fig.~\ref{fig:simulation_visualization_2D} shows a visualization
of the 2-D simulation environment and Fig.~\ref{fig:simulation_visualization_3D}
of the 3-D simulation environment.

The simulation experiments are designed to answer the
following questions: (1) Does the robot learn to circumvent unknown
obstacles and go to the goal position? (2) Is the robot able to plan
paths through narrow spaces? We use environments with a varying number
and size of obstacles. A large number of obstacles creates narrow spaces.

\begin{figure*}[thbp]
	\centering \vspace{0.5em}
	\begin{tabular}{ccc}
		\hspace{-0.5em}\includegraphics[width=0.30\textwidth]{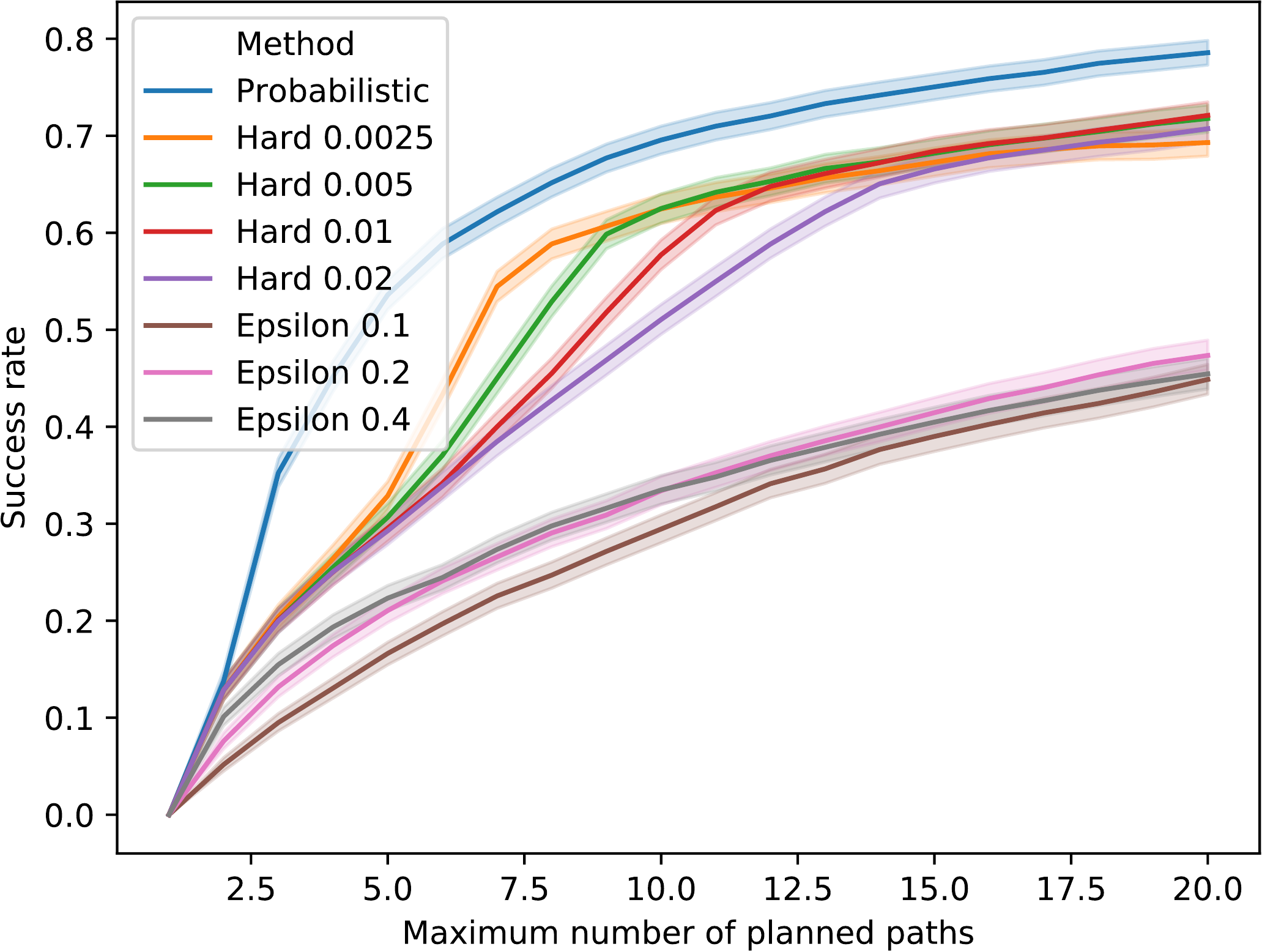} &
		\hspace{-1em}\includegraphics[width=0.30\textwidth]{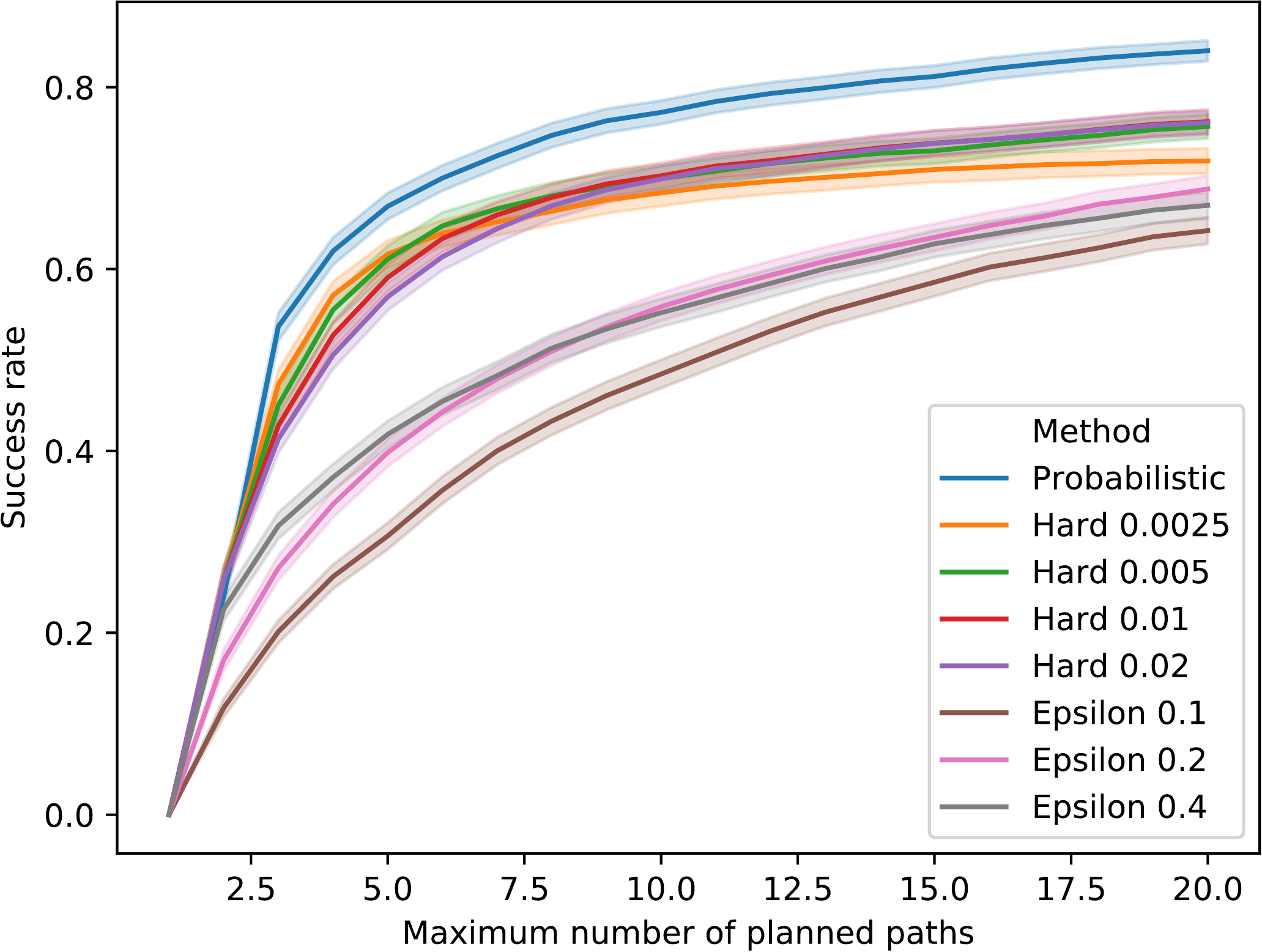} &
		\hspace{-1em}\includegraphics[width=0.30\textwidth]{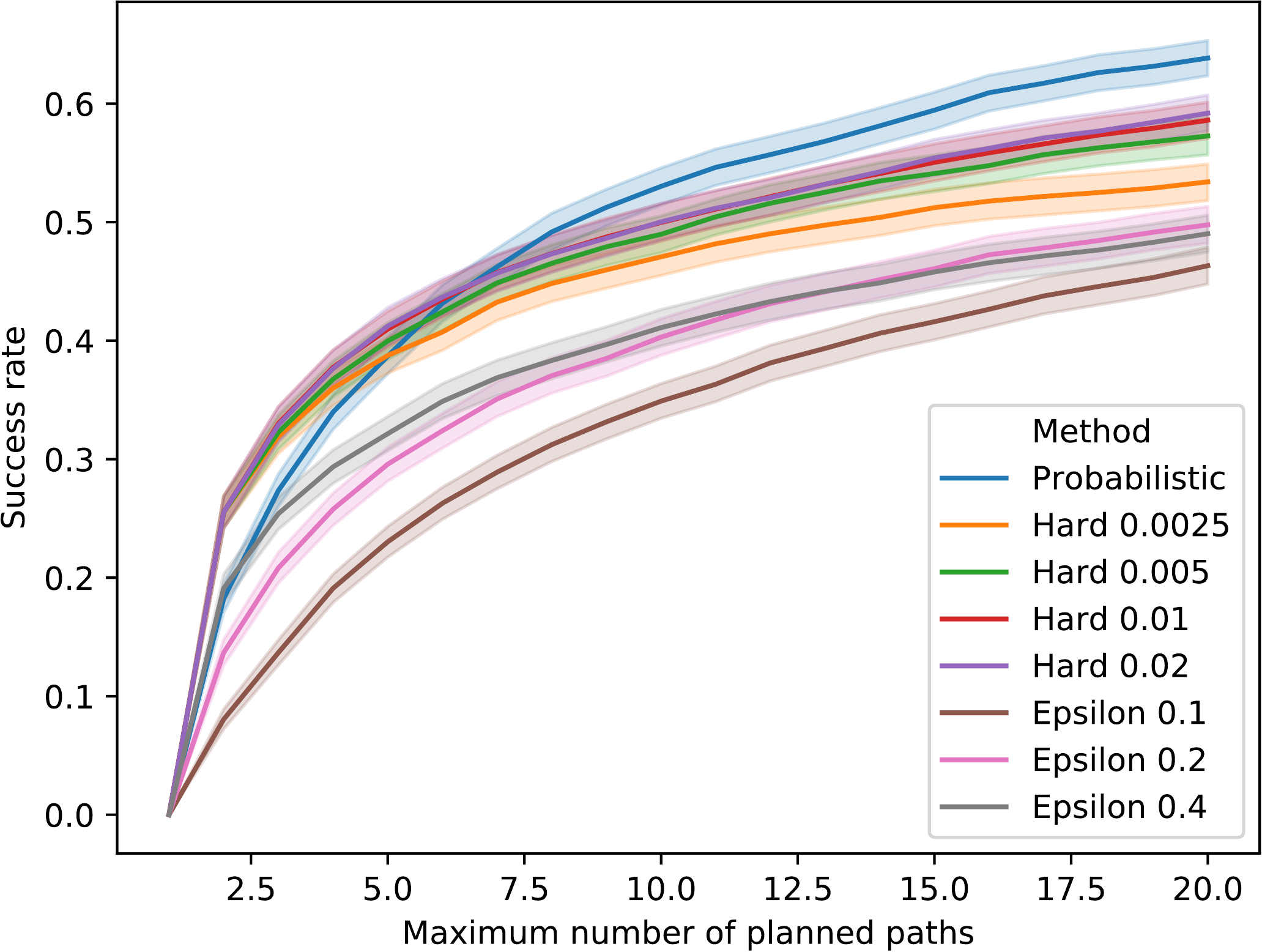} \\
		(a) 2-D & (b) 3-D & (c) 3-D with orientation
	\end{tabular}
	\caption{Evaluation in simulation. The results show for different maximum numbers of planned paths average success rate with a bootstrapped 95\% confidence interval. We ran experiments for a varying number of obstacles. For each obstacle number we performed 100 random evaluations and report the total numbers.
		\textbf{(a)} Planar 2-D robot moving from initial joint configuration to goal configuration with 1, 10, 30, and 50 disc obstacles. 
		\textbf{(b)} 3-D robot without taking orientation of the end-effector into account with 1, 10, 30, and 50 ball or disc obstacles.
		\textbf{(c)} 3-D robot taking orientation of the end-effector into account with 30, 50, 70, and 90 ball or disc obstacles.
		The proposed ``Probabilistic'' approach outperforms comparison methods.}
	\label{fig:simulation_results}
        \vspace{-0.2em}
\end{figure*}

\begin{figure*}[thbp]
	\centering
	\begin{tabular}{ccc}
		\hspace{-0.5em}\includegraphics[width=0.30\textwidth]{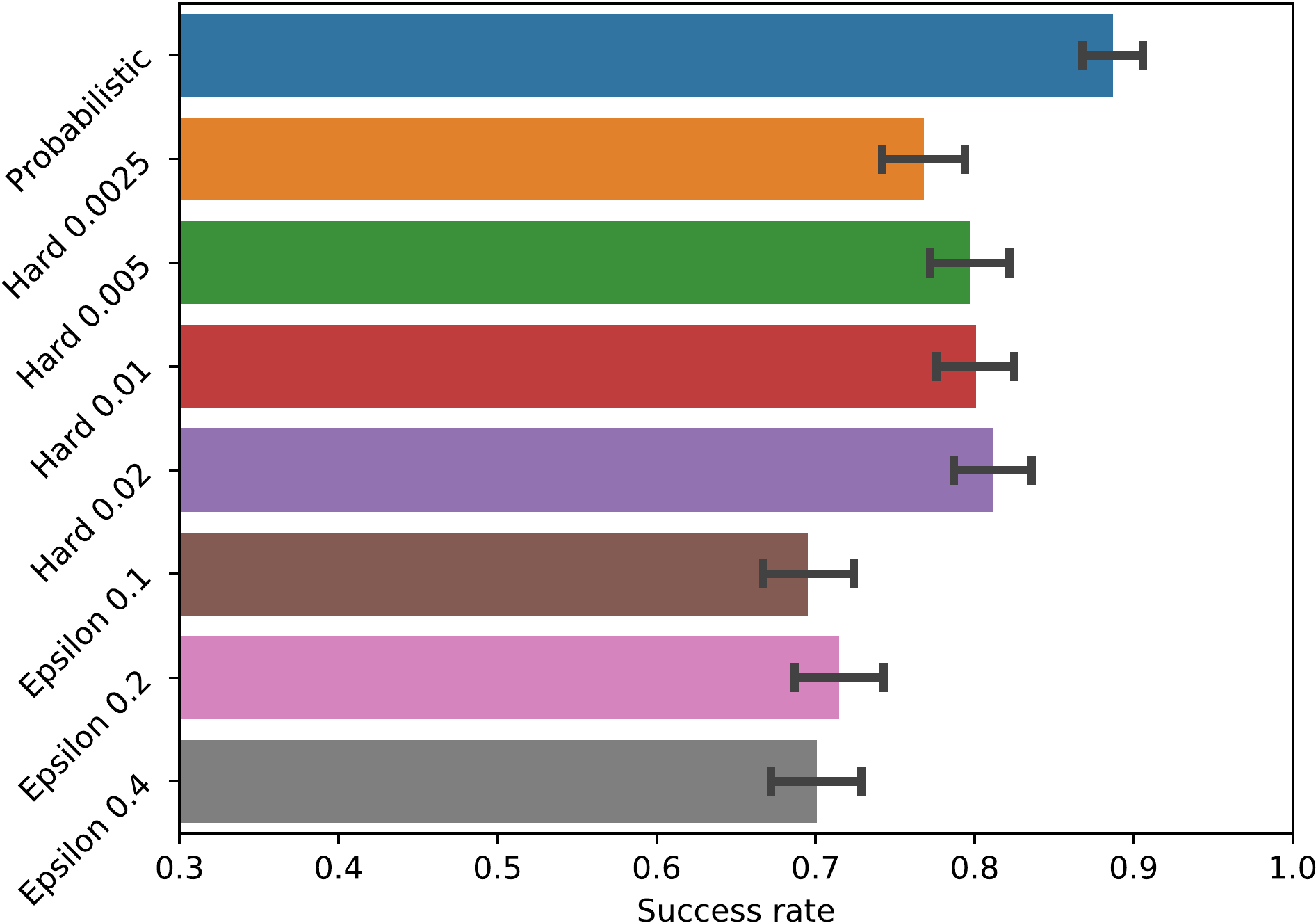} &
		\hspace{-1em}\includegraphics[width=0.30\textwidth]{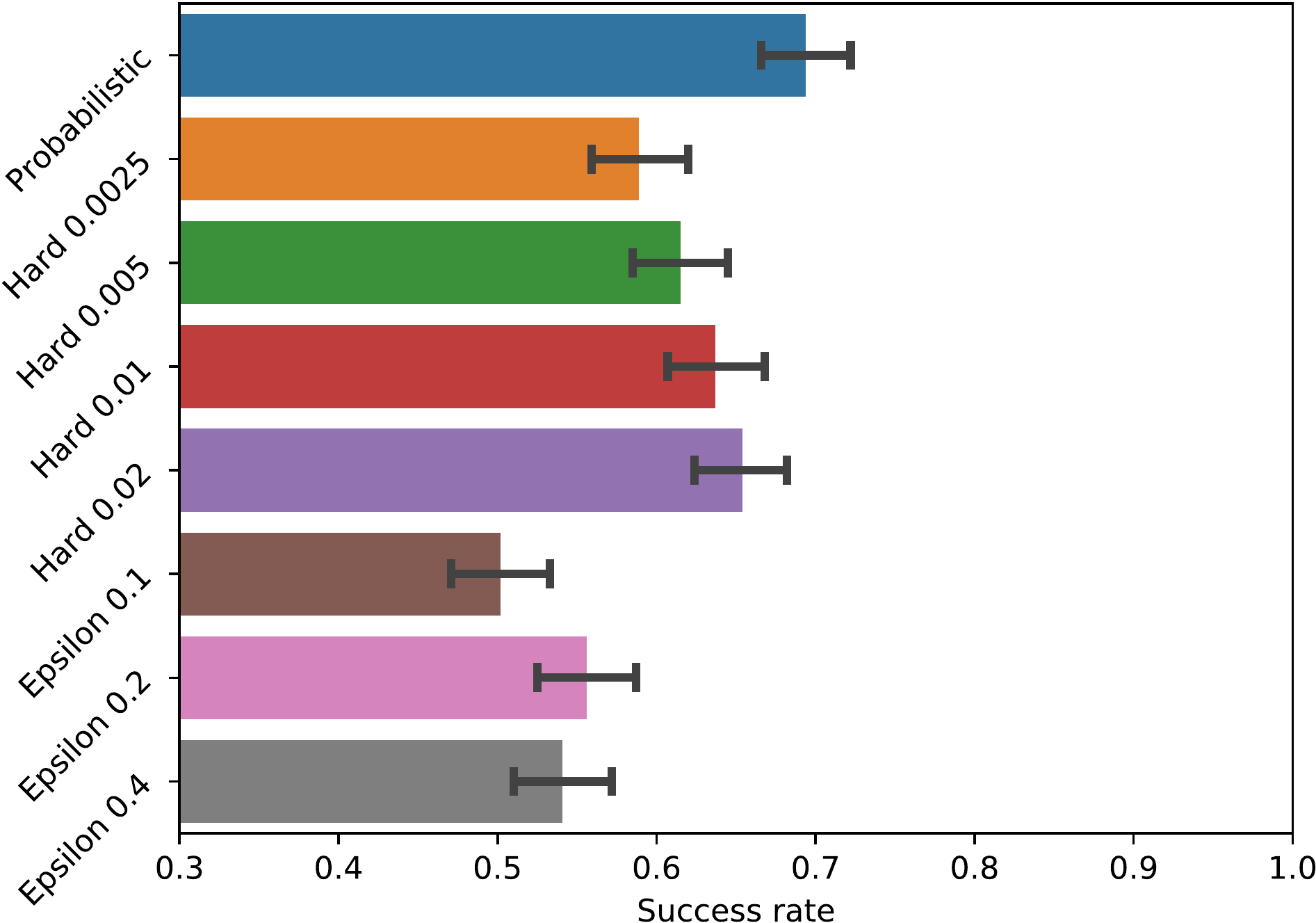} &
		\hspace{-1em}\includegraphics[width=0.30\textwidth]{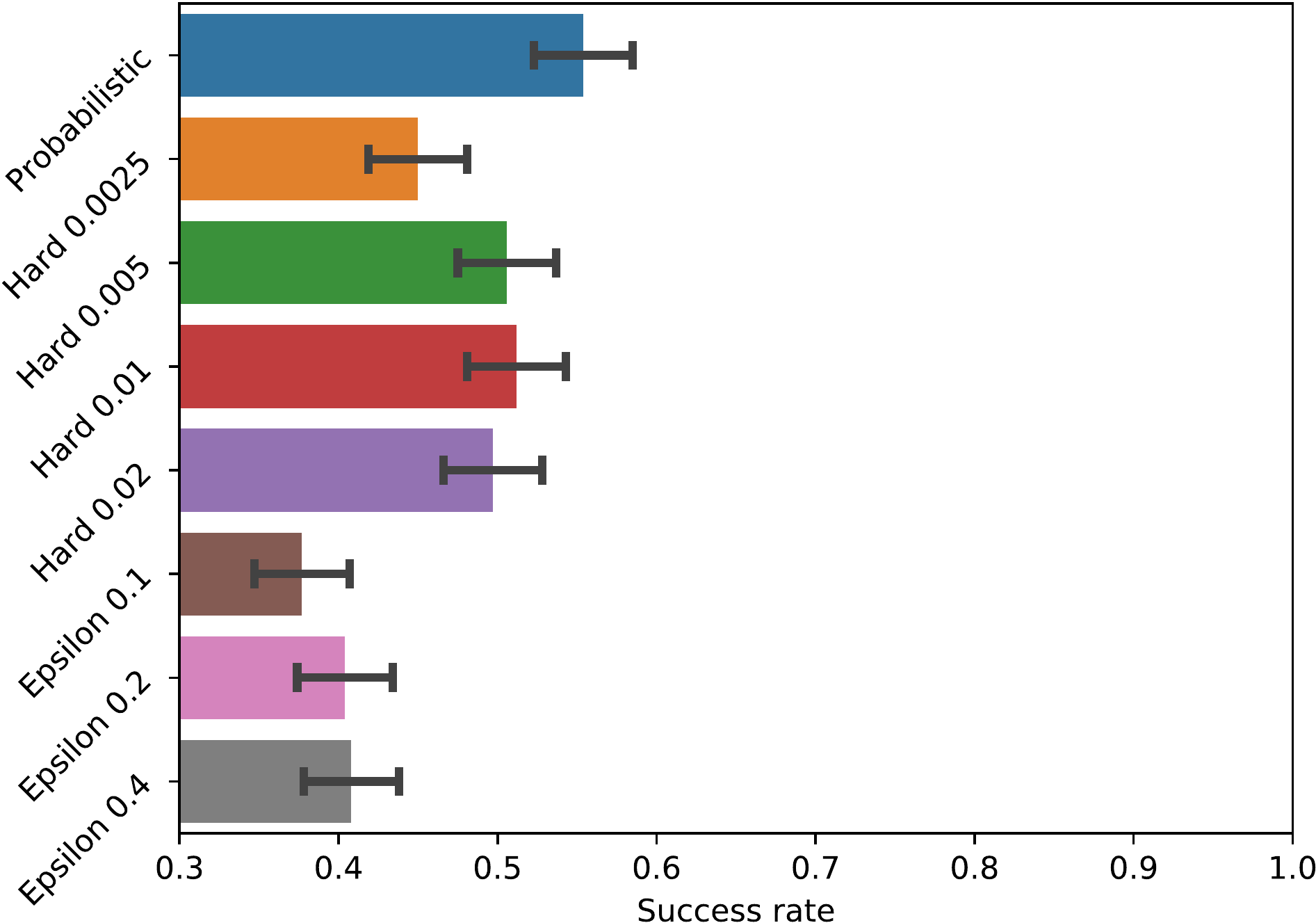} \\
		(a) 30 obstacles & (b) 50 obstacles & (c) 70 obstacles
	\end{tabular}
	\caption{Sensitivity analysis.
		For maximally $20$ planned paths in Fig.~\ref{fig:simulation_results}(c) we break down here the results according to the number of obstacles. While ``Probabilistic'' performs constantly well
		``Epsilon'' and ``Hard'' are sensitive to the choice of hyper-parameter value. With 30 obstacles ``Hard 0.02'' performs better than
		``Hard 0.01'' while with 70 obstacles ``Hard 0.01'' performs better than ``Hard 0.02''. This makes intuitively sense 
		since with more obstacles you have less space and do not want to block movements as much.
		With 30 obstacles ``Epsilon 0.2'' performs
		better than ``Epsilon 0.4'' while with 70 obstacles ``Epsilon 0.4'' performs better than ``Epsilon 0.2''. 
		An important thing to note is that in real robotic tasks the complexity or number and shape of obstacles is 
		usually not known
		beforehand and thus hyper-parameter tuning is not feasible.}
	\label{fig:simulation_sensitivity}
        \vspace{-0.2em}
\end{figure*}

Our approach does not employ hard constraints. 
Since we do not know the exact shape or location of obstacles hard
constraints could prevent the robot from going close to an obstacle,
and, the robot would potentially not be able to solve some
entanglements. We choose as comparison methods ``Hard'' and ``Epsilon''. ``Hard''
is an RRT version which is identical to our proposed approach except 
that we use a probability threshold to make failure probabilities
deterministic. This yields an algorithm called ``Hard'' similar to common
hard constraint RRT approaches. ``Hard'' allows for a fair
comparison to our proposed approach ``Probabilistic'' since the hard constraints
have similar shape to the probability distributions used by ``Probabilistic''.
``Epsilon'' is based on the
epsilon-greedy strategy~\cite{sutton2018reinforcement} widely used in reinforcement learning where
the robot moves directly towards the goal with epsilon probability and otherwise in a random
direction. We run different versions of these approaches with different hyper-parameter choices
denoted by ``Epsilon X'', where X denotes the epsilon value and ``Hard Y'' where Y denotes the
probability threshold value. Fig.~\ref{fig:simulation_results} shows the simulation results and Fig.~\ref{fig:simulation_sensitivity} shows
sensitivity analysis: ``Probabilistic'' outperforms the comparison methods in all environments and 
is not as sensitive to hyper-parameter choice.

\subsection{Robot experiments}
\begin{figure*}[thbp]
  \centering
  \begin{tabular}{cccc}
      \hspace{-0.6em}\includegraphics[width=0.245\textwidth]{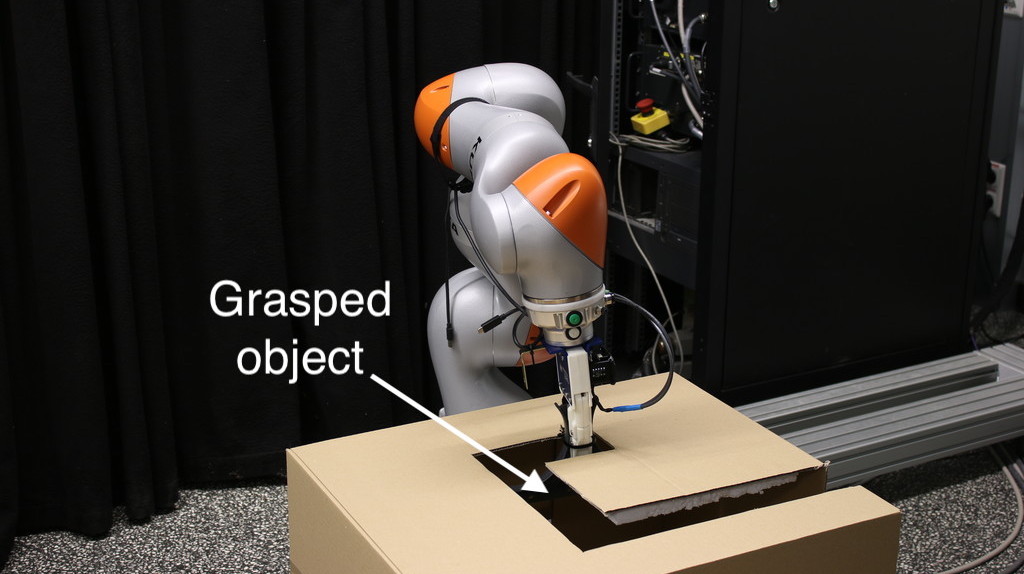}&
      \hspace{-0.95em}\includegraphics[width=0.245\textwidth]{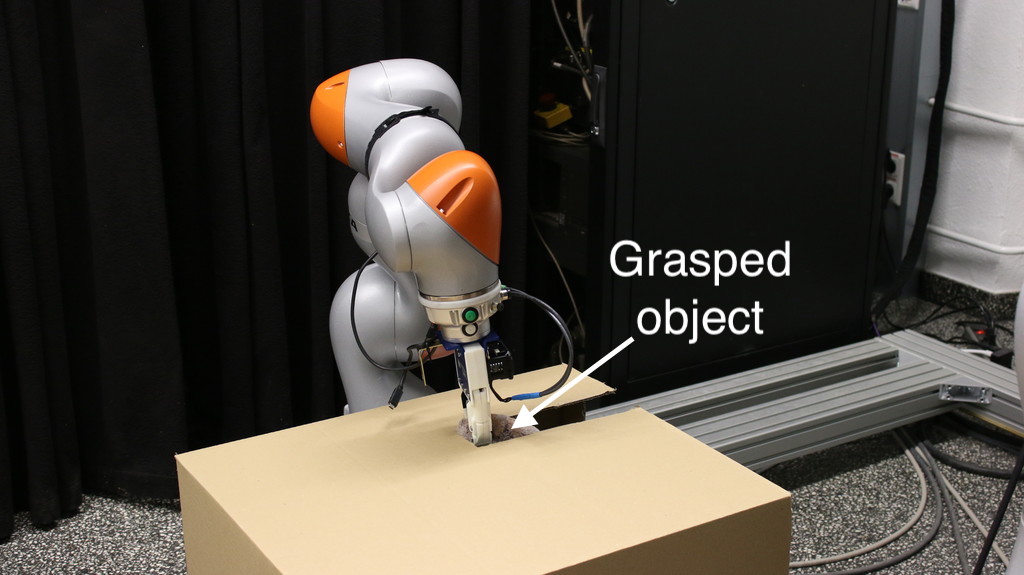}&
      \hspace{-0.95em}\includegraphics[width=0.245\textwidth]{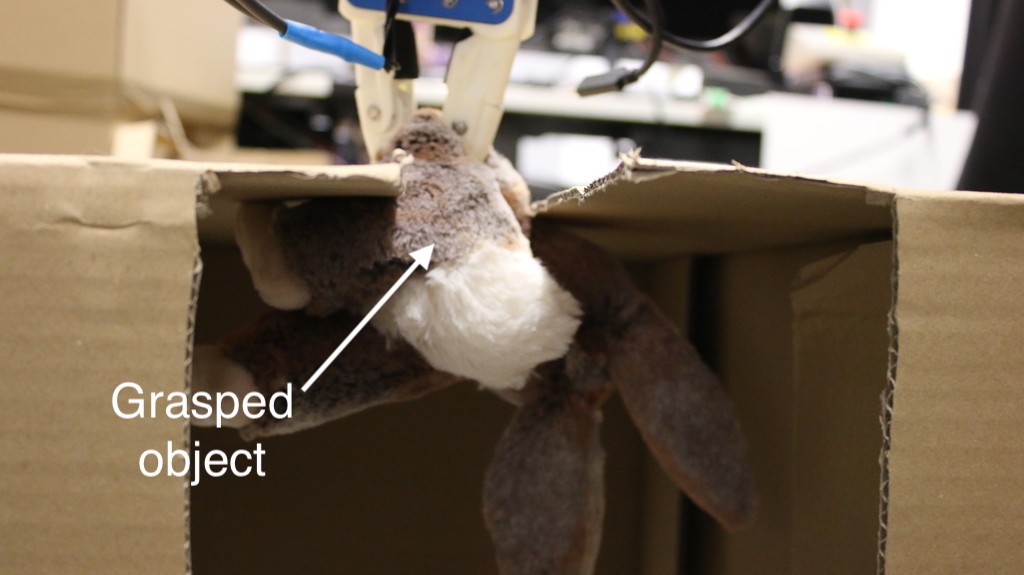}&
      \hspace{-0.95em}\includegraphics[width=0.245\textwidth]{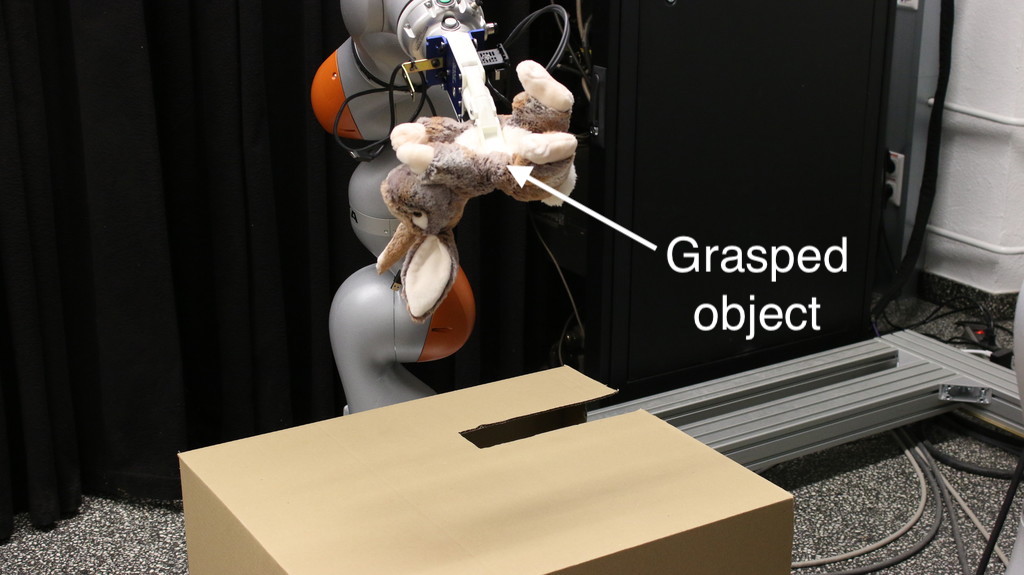}\\
      (a) Maze &
      (b) Bunny in a box & (c) Bunny in a box & (d) Bunny in a box
  \end{tabular}
  \caption{We have three robotic evaluation
    scenarios. \textbf{(Waste)} The waste evaluation scenario is
    shown in Fig.~\ref{fig:overview}(top) where a grasped gray plastic
    object is entangled in plastic and iron waste and needs to be
    moved away from the waste. When moving the object up flexible
    waste is pulled up requiring a specific combination of movement
    and rotation of the end-effector. \textbf{(Maze)} Gray plastic
    object needs to be moved out of a simple maze. This can be
    accomplished by either rotating the object at a suitable location
    and pulling it up or by moving it through the maze and out of the
    side of the cardboard box. \textbf{(Bunny)} Bunny needs to be
    moved out of the box through the hole on the side of the cardboard
    box: \textbf{(b)} initial configuration, \textbf{(c)} side view
    highlighting the difficulty of the task, \textbf{(d)} successful
    final configuration. The robot does not use any sensors, does not
    know anything about the objects or the environment, and gets
    feedback only about its own position. Similarly to simulation
    experiments, in order to disentangle the end-effector the robot
    tries movement paths until succeeding or until 20 paths have been
    tried. In ``Waste'' the environment is flexible and in ``Bunny''
    the grasped object is flexible.}  \label{fig:robot_experiments}
\end{figure*}

In the robot experiments, the proposed algorithm plans in 7
dimensional joint configuration space. We used a 7-DOF KUKA LBR robot
arm equipped with a SAKE gripper. The attached vision sensor shown in
Fig.~\ref{fig:overview} and Fig.~\ref{fig:robot_experiments} was not used in
the experiments. The robot experiments are designed to answer the
following questions: (1) Can the proposed approach be used with a real
robot? (2) Does the approach work in situations where the grasped
object is flexible or the environment is flexible? (3) How does the
proposed approach compare to the baselines in real robotic experiments?

\begin{table}[bhtp]
	\caption{How many times in ten evaluation runs each method
		succeeded to move the end-effector to the goal position in each robotic evaluation scenario shown in Figures \ref{fig:overview} and \ref{fig:robot_experiments}.}
	\label{tab:robot_performance}
	\begin{center}
		\begin{tabular}{l|ccc}
			\toprule
			Scenario & Epsilon & Hard & Probabilistic \\
			\midrule
			Waste & 6 & 3 & 7 \\
			Maze & 4 & 0 & 6 \\
			Bunny & 0 & 1 & 5 \\
			Total & 10 & 4 & 18 \\
			\bottomrule
		\end{tabular}
	\end{center}
\end{table}

Fig.~\ref{fig:robot_experiments} shows the three different real robot evaluation
scenarios. We chose scenarios containing
both flexible grasped and in-scene objects
which are common in applications such as waste segregation.
Table~\ref{tab:robot_performance} shows performance in
these scenarios.
``Probabilistic'' was best in all scenarios,
especially in the ``Bunny'' experiment which we considered the
hardest task a priori due to the constrained movements as
shown in Fig.~\ref{fig:robot_experiments}. ``Epsilon''
outperformed ``Hard'' in two out of three scenes which could be due
to the hyper-parameter sensitivity shown in the simulations in Fig.~\ref{fig:simulation_sensitivity}. In successful runs, the average number of executed robot paths was
$12.1$ for ``Probabilistic'', $9.3$ for ``Hard'', and $13.4$ for
``Epsilon''. Planning a path, on a 4-core Intel i7 CPU,
``Probabilistic'' took from $9.9$ seconds ($|\mathcal{M}| = 0$) to
$17.4$ seconds (($|\mathcal{M}| = 19$). ``Hard'' took
from $9.7$ seconds to $17.7$ seconds. ``Epsilon'' took less
than $0.2$ seconds for a planned path. Forward kinematics computation
was most time-consuming. We used a Python implementation except we used C for
forward kinematics. We expect speed ups with a full C-implementation
and when using data structures such as KD-trees.

\section{Conclusion}
When physically disentangling entangled objects we often do not have
object models, and, especially with cluttered irregularly shaped
objects, the robot can not create a model of the scene due to
occlusion. Based on purely joint configuration information our
approach learns a probability map of movement success and plans paths
based on path success probability. The approach sequentially executes
a planned path, incorporates information about movement success into
the probability map and plans a new path. We demonstrate the approach
successfully in simulated 2-D and 3-D environments and in three tasks with
a real 7-DOF KUKA LBR robot arm while outperforming comparison methods.
While the proposed approach worked well in the tasks we tried,
we are currently working on a POMDP solution allowing
more complex information gathering in even more complicated tasks.
Due to our probabilistic formulation incorporating prior knowledge
from sensors should be straightforward in future work.
To speed up the replanning part of our approach further, we could combine
ideas
from~\cite{bruce2002real,ferguson2006anytime,ferguson2006replanning}.
In particular, we could use caching together with discarding parts of the RRT tree
that are sub-optimal according to an admissible heuristic while
updating collision probabilities of existing RRT edges and nodes.
Finally, another possible direction for future work would be to adapt
our approach to new disassembly applications in recycling.
\bibliographystyle{IEEEtran}
\bibliography{root}

\end{document}